\pdfoutput=1

\documentclass[11pt]{article}

\usepackage[]{acl}

\usepackage{amsmath}
\usepackage{amsfonts}
\usepackage{booktabs}
\usepackage{times}
\usepackage{latexsym}
\usepackage{enumitem}
\usepackage{graphicx}
\usepackage{multirow}
\usepackage{subcaption}

\usepackage[T1]{fontenc}

\usepackage[utf8]{inputenc}

\usepackage{microtype}

\usepackage{inconsolata}

\usepackage{algorithm}
\usepackage{algpseudocode}
\usepackage{float}
\usepackage{tcolorbox}

\title{\textit{Threads of Subtlety}: Detecting Machine-Generated Texts \\ Through Discourse Motifs}

\author{Zae Myung Kim$^{1}$ \and Kwang Hee Lee$^{2}$ \and Preston Zhu$^{1}$ \and Vipul Raheja$^{3}$ \and Dongyeop Kang$^{1}$ \\
        University of Minnesota Twin Cities$^{1}$, Kumoh National Institute of Technology$^{2}$, Grammarly$^{3}$\\ 
        \texttt{\{kim01756,zhu00604,dongyeop\}@umn.edu}, \texttt{kwanghee@kumoh.ac.kr}, \texttt{vipul.raheja@grammarly.com}
}

\begin{document}
\maketitle

\begin{abstract}

With the advent of large language models (LLM), the line between human-crafted and machine-generated texts has become increasingly blurred. 
This paper delves into the inquiry of identifying discernible and unique linguistic properties in texts that were written by humans, particularly uncovering the underlying discourse structures of texts beyond their surface structures.
Introducing a novel methodology, we leverage hierarchical parse trees and recursive hypergraphs to unveil distinctive discourse patterns in texts produced by both LLMs and humans.
Empirical findings demonstrate that, although both LLMs and humans generate distinct discourse patterns influenced by specific domains, human-written texts exhibit more structural variability, reflecting the nuanced nature of human writing in different domains. 
Notably, incorporating hierarchical discourse features enhances binary classifiers' overall performance in distinguishing between human-written and machine-generated texts, even on out-of-distribution and paraphrased samples.
This underscores the significance of incorporating hierarchical discourse features in the analysis of text patterns.
The code and dataset are available at \url{https://github.com/minnesotanlp/threads-of-subtlety}.
\end{abstract}

\section{Introduction}\label{sec:introduction}

The emergence of powerful instruction-tuned large language models (LLMs) \citep{Ouyang2022Instruct,muennighoff-etal-2023-crosslingual, Kopf2023OpenAssistantC} has led to an explosion of machine-generated texts in both offline and online domains. Consequently, discerning the authorship of texts has become a significant challenge, spanning from educational settings to the landscape of online advertising \citep{Extance2023,DALALAH2023100822,GobAndrzejak2023}. Indeed, many efforts have been made to tackle this issue by constructing corpora of machine-generated and human-authored texts \citep{dou-etal-2022-gpt,guo2023close,li2023deepfake} and developing models and benchmarks to tell them apart \citep{wu2023survey,verma2023ghostbuster,su-etal-2023-detectllm,chakraborty-etal-2023-counter}. The consensus seems to be that while classifiers that make use of the presence of LLM-specific signatures in the generated texts perform relatively well on in-domain texts, their accuracy drops significantly with out-of-domain samples. Furthermore, these detectors can be fooled easily with ``paraphrasing attacks'' even with in-domain samples \citep{sadasivan2023aigenerated,krishna2023paraphrasing}.

\begin{figure}[t!]
  \includegraphics[width=\columnwidth]{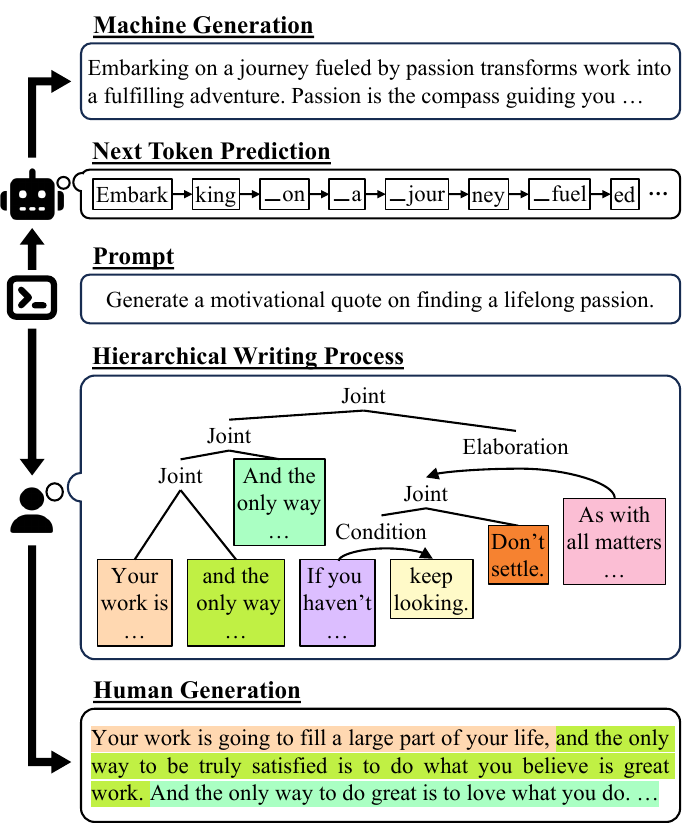}
  \caption{Human writers often employ hierarchical linguistic structures in writing whereas LLMs primarily operate by the sequential next token prediction task.
  \vspace{-4mm}}
  \label{fig:conceptual_fig}
\end{figure}

\begin{figure*}[!ht]
  \includegraphics[width=\textwidth]{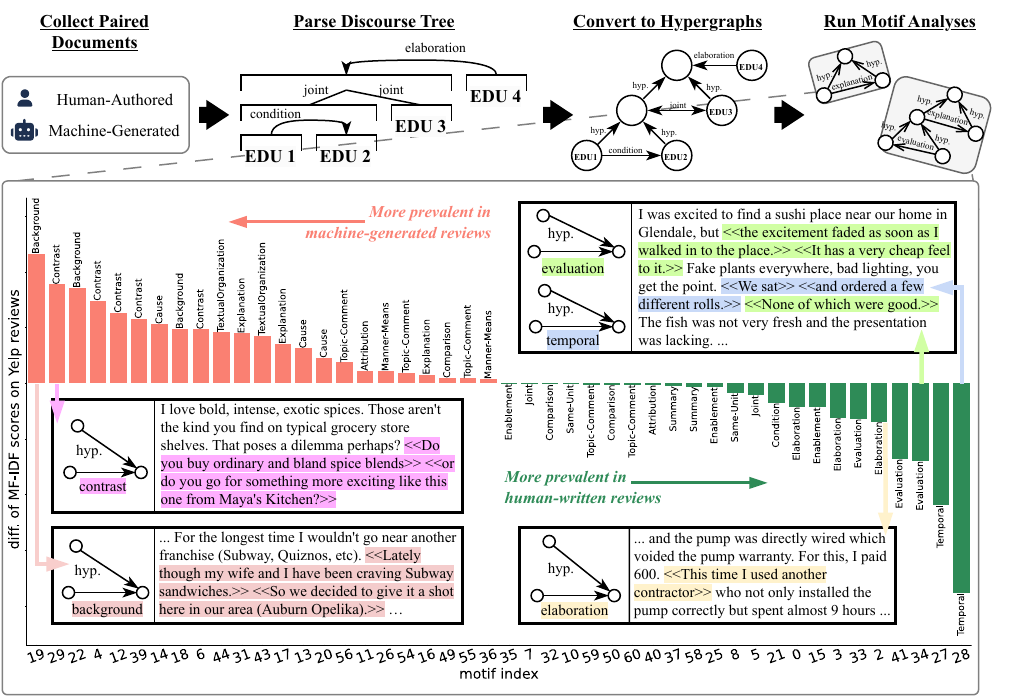}
  \caption{Difference in motif distribution of machine-generated and human-written texts for Yelp domain. Below are the top discourse motifs for each authorship and their corresponding discourse relations with examples. Text enclosed in angle brackets denotes the EDUs involved in the relation.}
  \label{fig:main_intro_fig}
\end{figure*}

This raises interesting questions on the underlying nature of human-written texts: ``Are there any discernible, unique properties within texts crafted by humans?'' and if so, ``Might these distinctive signatures manifest at levels beyond surface structure?'' Undeniably, how we write varies greatly depending on the domain and intent addressed by the texts. Consider, for instance, the distinction between academic writing and a casual response to a Reddit post.
The former pays much more attention to the logical progression of arguments and overall structural coherence, while the latter is characterized by a more spontaneous and less structured thought process.

On the contrary, LLMs, in consideration, are autoregressive models that generate the next token based on the previous sequence, i.e. $P(x_{t} | x_{1}..x_{t-1})$, without explicitly modeling the hierarchical structures in the process.
Figure \ref{fig:conceptual_fig} illustrates this structural difference in the writing processes of LLMs and humans. 
We note that LLMs may also internally capture these hierarchical structures to some degree as a by-product, especially in their self-attention matrices \citep{xiao-etal-2021-predicting,huber-carenini-2022-towards}.
However, our interest lies in exploring whether there exist distinctive hierarchical structures that can aid in distinguishing their authorship.\footnote{We provide a more comprehensive comparison with prior studies in Appendix \ref{sec:related_work}.}

To answer these inquiries systematically, we draw inspiration from discourse analysis in linguistics \citep{mann1987rhetorical}, which uncovers structures found within a sentence, between sentences, and among the paragraphs of a document. Specifically, Figure \ref{fig:main_intro_fig} highlights our main approach:
Given texts written by humans and LLMs, we construct a hierarchical parse tree for each document. Subsequently, we transform these trees into recursive hypergraphs, allowing us to perform network motif analysis of discourse relationships. The essence of our analysis lies in computing the difference in the distribution of these \textit{``discourse motifs''} between texts generated by machines and those crafted by humans, identifying discernible discourse patterns within each authorship. 


\begin{figure*}[!t]
\centering
{\includegraphics[width=\textwidth]{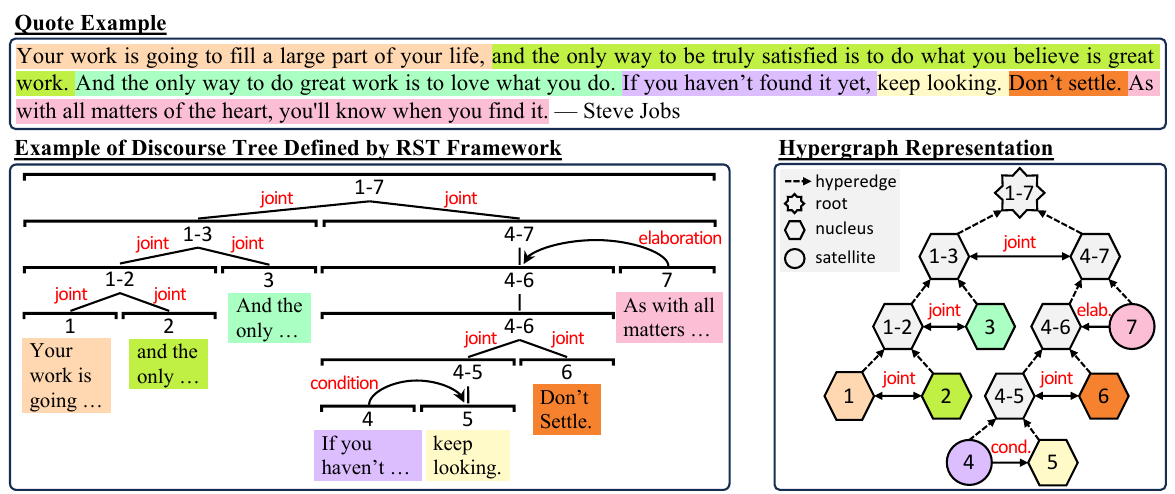}}
\caption{A quote from Steve Jobs and its RST tree converted into a hypergraph form. A hexagonal node represents the ``nucleus'' node, while a circular one denotes the ``satellite'' node. Each node is labeled with a span of EDU indices that it covers. The star-shaped node is the root node of the graph, encompassing all subgraphs and EDUs.\vspace{-4mm}}
\label{fig:hypergraph_example}
\end{figure*}

We summarize our contributions as follows:
\begin{itemize}[noitemsep]
    \item We demonstrate that hierarchical discourse structures, as defined by RST, can be effectively modeled using recursive hypergraphs.
    \item To the best of our knowledge, we are the first to demonstrate that integrating hierarchical discourse-level features into the authorship detection task results in enhanced and more robust performance.
    \item Our empirical findings suggest that the robustness against paraphrasing attacks stems from the preservation of higher-level discourse structures, despite significant variations at the sentence level.
    \item Addressing a critical gap in existing research, where machine-generated texts typically fall within the range of 200 to 500 tokens, we construct a dataset specific to the creative writing domain. This dataset features 20 stories with lengths of up to 8K tokens, providing a more comprehensive perspective on longer-form machine-generated content.
\end{itemize}

\section{Modeling Hierarchical Structures}\label{sec:hierarchical_structures}
To explore unique patterns that may extend beyond surface-level structure, it is imperative to employ an expressive framework capable of representing hierarchical structures within texts. Human writers strategically utilize textual structures to systematically convey meaning, thereby augmenting the clarity, coherence, and persuasiveness of their written compositions. In the field of linguistics, the study of document structure is within the domain of discourse frameworks \citep{mann1987rhetorical,miltsakaki-etal-2004-penn,lascarides2007segmented}. This paper adapts Rhetorical Structure Theory (RST) \citep{mann1987rhetorical} due to its widespread acceptance and the availability of openly accessible pre-trained parsers. In this paper, we make use of the \textsc{DMRST} parser released by \citet{liu-etal-2021-dmrst}.

\subsection{Rhetorical Structure Theory}
RST assumes that any well-written document can be parsed into a (recursive) discourse tree. In this tree, each leaf node, called an Elementary Discourse Unit (EDU), corresponds to a phrase within a sentence, while higher-level nodes integrate these lower-level representations into more complex structures, such as phrases to sentences to paragraphs. It does so by assigning a discourse label (e.g., elaboration, contrast, cause, effect, etc.) to a relation (i.e., edge) between nodes. The linked nodes can be either ``nuclei'' (main ideas) or ``satellites'' (supporting details) depending on their relative importance in the relation. Figure \ref{fig:hypergraph_example} illustrates how a quote from Steve Jobs is segmented into seven EDUs, each marked in a distinct color, and parsed to form a discourse tree structure.

\subsection{Conversion to Hypergraphs}
Within the RST framework, discourse relations emerge not only through the linking of EDUs but also by encapsulating other relations. Specifically, RST relations are applied to a text ``recursively'' until every unit within the text becomes a constituent of an RST relation. 
This recursive application implies that a single EDU can form an RST relation with another individual EDU, and also become part of a relation involving a group of EDUs.
For example, in Figure \ref{fig:hypergraph_example}, EDU 4 and EDU 5 are linked by a ``condition'' relation, while EDU 6 forms a ``joint'' relation with the collective unit of EDU 4 and EDU 5. Additionally, since the ``joint'' relation encompasses the ``condition'' relation, RST trees inherently demonstrate recursive encapsulations.

This recursive encapsulation characteristic of RST trees naturally corresponds with recursive hypergraphs, also referred to as ``Ubergraphs'' in mathematics \citep{Joslyn2017UbergraphsAD}, where a discourse relation aligns with a \textit{recursive hyperedge}. 
In short, hypergraphs generalize traditional graphs by allowing edges (or ``hyperedges'') to connect more than two nodes. The recursive hypergraphs further generalize the standard hypergraphs by allowing hyperedges to contain not only nodes but also other hyperedges. This leads to a hierarchical structure, where hyperedges at different levels of recursion represent relationships at varying levels of abstraction.

To facilitate ease of analysis, the hypergraph can be transformed into an isomorphic Levi graph (i.e., a standard traditional graph form) by introducing a dummy edge (depicted as a dotted line in Figure \ref{fig:hypergraph_example}) that signifies the expanded hyperedge relation. By expanding the nested recursions, we are now able to run standard network analysis algorithms as well as train Graph Attention Networks (GATs) \citep{velickovic2018graph}, leveraging the inherent discourse structure and semantics of documents.

We note that although \citet{NEURIPS2020_217eedd1} briefly explored the transformation of text into recursive hypergraphs, their methodology defined hyperedge relations as the relationships between entities (such as persons, places, etc.) occurring within the sentence-level organization of the text.

\section{Proposed Method: Revealing Key Patterns via \textit{Discourse Motifs}}\label{sec:revealing_key_patterns}
This paper focuses on finding distinctive structural patterns, namely, hierarchical discourse relations based on the RST framework (\S \ref{sec:hierarchical_structures}). Our main investigative approach involves analyzing the distribution of discourse motifs, which are \textit{recurring and statistically significant subgraph patterns} within a larger network. These motifs are often considered as the building blocks of larger networks and can provide insights into the organization and dynamics of the system \citep{milo2002networkmotifs,Takes2018MultiplexNM}. 
As illustrated in Figure \ref{fig:main_intro_fig}, our proposed approach involves gathering documents written by humans and LLMs, parsing their RST trees, transforming them into hypergraphs, and running motif analyses to understand how individuals (both humans and machines) craft texts. The following subsections illustrate how the discourse motifs are systematically formed (\S\ref{subsec:discourse_motifs}) and scored using our proposed metrics (\S\ref{subsec:metrics}).

\subsection{Discourse Motifs as Unions of Triads}\label{subsec:discourse_motifs}
One of the standard ways to construct motifs is to generate all possible non-isomorphic subgraphs of a fixed number of vertices. However, as the number of vertices increases, the number of possible patterns increases super-exponentially, rendering their counting extremely inefficient.
Complicating matters further, our RST graphs are \textit{directed} with their edges \textit{labeled} with discourse relations, which further increases the complexity.\footnote{It is worth noting that some discourse relations can be bidirectional as well.}

With this in mind, we start by generating all possible motifs with 3 vertices, i.e., ``triads.'' This is a minimal form of motifs other than the single edge case, ``dyad.'' We select the triads to be the basis motifs as we find that when converting a recursive RST tree into a standard graph, it always forms \textit{the union of subgraphs with 3 vertices}. We highlight the proof in Appendix \ref{sec:app:unions}.

This process results in 69 non-isomorphic motifs. To create larger motifs, we pick two triads from all possible pairs, join them at all possible nodes, and only keep the non-isomorphic ones, resulting in unions of two triads. We term these configurations as ``double-triads.'' We conduct this process one more round with pairs of a (single-)triad and a double-triad to produce a ``triple-triad.''\footnote{As we are joining motifs at their nodes, the total number of nodes for double-triads and triple-triads will be smaller than 6 and 9, respectively.}
We note that depending on how these multiples of triads are joined together, we can form discourse motifs with varying numbers of vertices such as 4, 5, and 7. Table \ref{tab:three_motifs_sizes} shows the number of identified motifs for each type. Graphical examples of motifs can be found in Appendix \ref{sec:app:motif_graphs}.

\begin{table}[!h]
    \centering
    \resizebox{0.8\columnwidth}{!}{
    \begin{tabular}{c|c|c}
        \toprule
        \textbf{Single-Triads} & \textbf{Double-Triads} & \textbf{Triple-Triads} \\
        \midrule
        69 & 592 & 2,394\\
        \bottomrule
    \end{tabular}}
    \caption{Number of non-isomorphic discourse motifs.}
    \label{tab:three_motifs_sizes}
\end{table}

\subsection{Metrics for Analyses}\label{subsec:metrics}

Once we gather potentially useful discourse motifs, we simply count the number of isomorphic subgraphs for each motif across all documents in the datasets. However, as the graph isomorphism test is (possibly) NP-intermediate, the process requires heavy engineering feats to maximize computational efficiency, such as hashing the subgraphs \citep{ShervashidzeSLMB11} and multiprocessing the datasets.\footnote{The code and the processed datasets for the experiments can be found at \url{https://github.com/minnesotanlp/threads-of-subtlety}.}
Afterward, we can identify distinctive motifs by examining their difference distributions based on their normalized counts (i.e., frequencies) (\S\ref{subsubsec:motif_diff_dist}) as well as calculating our proposed metric termed ``motif frequency-inverse document frequency (MF-IDF)'' measure (\S\ref{subsubsec:mfidf}).
The rationale behind devising the MF-IDF scoring lies in pinpointing and prioritizing significant motifs, as counting the entire motif distribution is inefficient. An analogy can be drawn to crafting a localized tokenization scheme for LLM pre-training, wherein tokens represent motifs, and texts serve as hypergraphs in our context.

\subsubsection{Motif Difference Distributions}\label{subsubsec:motif_diff_dist}
As a simple measure, we consider the difference distribution between features of motifs present in graphs. Specifically, we utilize their \textit{motif frequencies}, $\mathcal{F}(\cdot)$ and \textit{weighted average depths}, $\mathcal{D}(\cdot)$ for (document) graphs generated by machines ($\mathbb{D}_{\text{machine}}$) or authored by humans ($\mathbb{D}_{\text{human}}$):
\begin{equation}
    \mathcal{F}(\mathbb{D}_{\text{diff}}) = \mathcal{F}(\mathbb{D}_{\text{machine}}) - \mathcal{F}(\mathbb{D}_{\text{human}})
\end{equation}
\begin{equation}
    \mathcal{D}(\mathbb{D}_{\text{diff}}) = \mathcal{D}(\mathbb{D}_{\text{machine}}) - \mathcal{D}(\mathbb{D}_{\text{human}})
\end{equation}
These distributions represent the discrepancies of motifs between the two classes of authorship and can be computed across different domains of texts as well.

\paragraph{Motif Frequency (MF).} For each motif $m$ in a set of identified motifs $M$, we compute motif frequency by counting its occurrences in a graph $g$ and normalize it by the total number of motifs in the graph. The frequency is then averaged over the corresponding dataset $\mathbb{D}$:
\begin{equation}
    \mathcal{F}(m, \mathbb{D}) = \frac{1}{|\mathbb{D}|} \sum_{g \in \mathbb{D}} \frac{\text{count}(m, g)}{\sum_{m' \in M} \text{count}(m', g)}
\end{equation}

\paragraph{Weighted Average Depth (WAD).} To better capture the hierarchical nature of discourse graphs, we also calculate the average depths of each motif $m$ appearing in a graph $g$.
Let $S(m,g)$ be a collection of subgraphs of a graph $g$ that are isomorphic to a motif $m$.
Then, the depth of a motif $m$ in a graph $g$ is measured by the \textit{mean position}
\begin{equation}
        \overline{d}(m, g) = \sum_{g' \in S(m,g)}\sum_{v \in V_{g'}}\frac{d(v, v_{root})}{|S(m,g)||V_{g'}|}
\end{equation}
 where $V_{g'}$ denotes the set of nodes of a subgraph $g'$ and $d(v, v_{root})$ denotes the distance from a node $v$ to the root $v_{root}$, calculated in $g$. 
 It is then weighted by the count of the corresponding motif and normalized by the total number of counts:
\begin{equation}
    \mathcal{D}(m, \mathbb{D}) = \frac{1}{|\mathbb{D}|} \sum_{g \in \mathbb{D}} \frac{\text{count}(m, g)\cdot\overline{d}(m, g)}{\sum_{m' \in M} \text{count}(m', g)}
\end{equation}
This effectively measures the average position of each discourse motif appearing in the graphs.

\subsubsection{Motif Frequency-Inverse Document Frequency (MF-IDF)}\label{subsubsec:mfidf}
Inspired by the term frequency-inverse document frequency (TF-IDF) measure from information retrieval domain \citep{sparckjones_statistical_1972}, we propose ``motif frequency-inverse document frequency'' (MF-IDF), noting that a motif ($m$) can be considered as a vocabulary of documents that are graphs ($g \in \mathbb{G}$) in our case:
\begin{align*}
    \text{MF-IDF}(m, g, \mathbb{G}) = \mathcal{F}(m, \{g\}) \cdot \text{IDF}(g, \mathbb{G})
\end{align*}
where:
\begin{align*}
    \mathcal{F}(m, \{g\}) = \frac{\text{count}(m, g)}{\sum_{m'\in M}\text{count}(m', g)} \\
    \text{IDF}(g, \mathbb{G}) = \log \frac{1 + |\mathbb{G}|}{1 + |g \in \mathbb{G}: m \in g)|} \\
\end{align*}
\section{Experimental Setup}\label{sec:exp_setup}
Following the generation of three sets of discourse motifs for single-, double-, and triple-triads (\S\ref{subsec:discourse_motifs}), we evaluate them using the MF-IDF metric and retain those with scores surpassing at least one standard deviation. This yields a total of 207 motifs, with a detailed breakdown presented in Table \ref{tab:three_selected_motifs_sizes}.

\begin{table}[!h]
    \centering
    \resizebox{0.8\columnwidth}{!}{
    \begin{tabular}{c|c|c}
        \toprule
        \textbf{Single-Triads} & \textbf{Double-Triads} & \textbf{Triple-Triads} \\
        \midrule
        31 & 96 & 80 \\
        \bottomrule
    \end{tabular}}
    \caption{Number of selected discourse motifs with MF-IDF scores exceeding the threshold of at least one standard deviation.}
    \label{tab:three_selected_motifs_sizes}
\end{table}

Section \ref{subsec:baseline} outlines the utilization of our proposed features (\S\ref{sec:revealing_key_patterns}) within both the baseline models and analyses; and Section \ref{subsec:datasets} describes the datasets designated and proposed for experimentation.

\subsection{Baseline Models}\label{subsec:baseline}

\begin{figure}[!th]
\centering
{\includegraphics[width=\columnwidth,trim={0.01cm 0.0cm 0.04cm 0.01cm},clip]{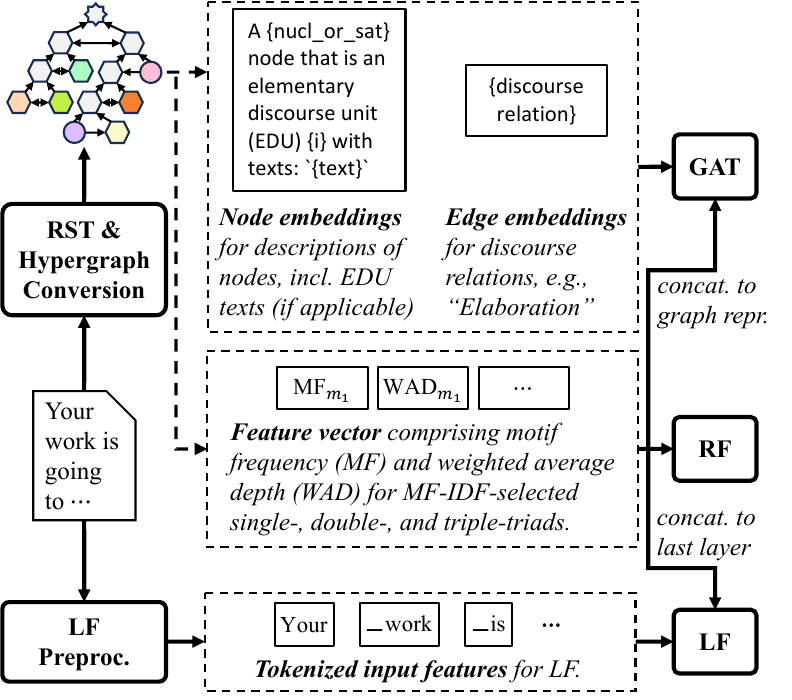}}
\caption{A data flow diagram illustrating how various types of input features are fed into the three baseline models: Graph Attention Network (GAT), Random Forest (RF), and Longformer (LF).\vspace{-5mm}}
\label{fig:pipeline}
\end{figure}

To evaluate the usefulness of discourse motifs, we integrate them into three baseline models for authorship detection (\S\ref{subsec:authorship_detection}): Random Forest (RF) \citep{breiman2001random}, Graph Attention Network (GAT) \citep{velickovic2018graph}, and Longformer (LF) \citep{beltagy2020longformer}.
Figure \ref{fig:pipeline} depicts an overall process of various types of input features being handled by the different models with varying numbers of model parameters and granularity of inputs.

RF is chosen to evaluate the baseline performance solely based on the discourse motif features (defined in \S\ref{subsubsec:motif_diff_dist}) without considering the semantic content of the texts.

GAT receives the EDU-level representation of the texts as initial node embeddings, along with the discourse edge labels as edge embeddings. The texts for both EDUs and discourse labels are embedded by using Sentence-BERT \citep{reimers-gurevych-2019-sentence} model, \textsc{all-MiniLM-L6-v2}. The discourse motif features are then appended to the mean-pooled representation of the graph.

LF is initialized with pretrained weights\footnote{\textsc{allenai/longformer-base-4096}, available on \url{https://huggingface.co}.} and fed with the entire sequence of texts, in addition to the discourse motif features concatenated to its [CLS] token.
Specifically, we align our use of LF with the methodology described by \citet{li2023deepfake}, which details a state-of-the-art detection model.
However, we introduce three key distinctions: (i) we train a single classifier across all ten domains of the \textsc{DeepfakeTextDetect} dataset, (ii) we utilize approximately 56\% of the original dataset to mitigate data imbalance issues, and (iii) we avoid adjusting the decision boundary across ten domain-specific classifiers. Our goal is to develop a general classifier capable of evaluating discourse motifs across various model architectures, rather than tailoring optimizations for individual domains.

We note that our approach can be seamlessly integrated into other baselines by concatenating discourse motif features with input representations. Additionally, it can enhance other detection strategies by providing extra statistical features for zero-shot detectors or by aiding in the selection of more challenging and deceptive samples for adversarial learning setups.


\paragraph{Formality Scorer.}
To identify a relationship between formality style of texts and discourse hierarchies (studied in \S\ref{subsec:formality_scorer}), we make use of a publicly available sentence-level formality scorer \citep{Babakov2023Formality} that is finetuned on a dataset for formality style transfer task, \textsc{GYAFC} \citep{rao-tetreault-2018-dear}. The model is based on a \textsc{RoBERTa-base} model \citep{zhang2020pegasus}.

\subsection{Datasets}\label{subsec:datasets}
Our experiments are conducted on two existing benchmark datasets and one new dataset (\textsc{TenPageStories}) that we created for more in-depth analysis.

\paragraph{\textsc{HC3-English}.} \citet{guo2023close} published the first corpus on human writer vs. ChatGPT comparison. It comprises 24K \textit{paired} responses from human authors and ChatGPT for mostly question-answering (QA) style prompts across 5 domains: Reddit-ELI5, medicine, finance, open QA, and Wikipedia-CS.

\paragraph{\textsc{DeepfakeTextDetect}.} \citet{li2023deepfake} proposed \textsc{DeepfakeTextDetect} dataset which consists of 448K \textit{mostly unpaired} human-written and machine-generated texts from 10 diverse datasets including news article writing, story generation, opinion writing, etc. The machine-generated texts are from 27 mainstream LLMs from 7 sources such as OpenAI and LLaMA. Out of ten domains in the dataset, we present results on the following five considering the diversity and sizes of domains: review writing (\textsc{yelp}), story writing (\textsc{wp}), argument writing (\textsc{cmv}), news summarization (\textsc{xsum}), and writing descriptions for scientific tables (\textsc{sci\_gen}).

\paragraph{\textsc{TenPageStories} (Ours).} Existing datasets for the authorship detection task primarily contain short documents of 200 to 500 tokens, limiting the capture of long-term discourse relations; to overcome this limitation, we construct exceptionally lengthy generations based on 20 fictional stories on Project Gutenberg published in early November 2023.\footnote{\url{https://www.gutenberg.org}}

Our generated content is prepared in three distinct settings:
\begin{enumerate}[noitemsep]
    \item \textbf{Unconstrained}: We provide the first human-written paragraph of a story and instruct the LLM to iteratively generate continuations of the story. However, due to the limited input length of the existing discourse parser, we did not use the unconstrained generations in experiments.
    \item \textbf{``Fill-in-the-gap''}: We mask $N \in {1, 3, 5}$ paragraphs between a preceding human-written paragraph and a subsequent human-written paragraph.
    \item \textbf{Constrained ``fill-in-the-gap''}: Similar to (2), but the masked paragraph(s) now include the first and last sentences of the corresponding human-written paragraph(s), providing more guided contexts.
\end{enumerate}

These settings are designed to enable the observation of long-term discourse patterns while varying the constraints applied to the original contexts. We generate content up to 8K tokens ($\approx$10 A4 pages) in an iterative manner, continuing from the previously generated texts. We provide more details on the dataset construction in Appendix \ref{sec:app:ten_page_stories}.


\section{Results}\label{sec:experimental_results}

We present findings from three sets of experiments:
\begin{enumerate}[noitemsep]
    \item Evaluation of the utility of discourse motifs in the authorship detection task including the ``paraphrase attack'' scenario (\S\ref{subsec:authorship_detection})
    \item Investigation of structural variations after paraphrasing (\S\ref{subsec:upper_vs_lower})
    \item Identifying the relationship between formality and hyperedges (\S\ref{subsec:formality_scorer})
\end{enumerate}

\subsection{Human vs. Machine Authorship Detection}\label{subsec:authorship_detection}
In this section, we report F1 scores for the binary classification task, following the experimental setup detailed in Section \ref{subsec:baseline}.

\paragraph{\textsc{HC3} and \textsc{DeepfakeTextDetect}.}
From Table \ref{tab:overall_f1}, we can see that incorporating motif information consistently enhances classification performance across various base encoders, underscoring the effectiveness of discourse motifs as auxiliary information for capturing deeper linguistic structures beyond surface lexicons. Specifically, when the GAT model leverages the overall hierarchical discourse structure, notable performance gains are observed (e.g., 0.67 $\rightarrow$ 0.73 on \textsc{HC3}). Additionally, given the LF model's proficiency in comprehending lengthy texts, the inclusion of discourse motifs offers explicit structural insights, further enhancing the performance. Even on the paraphrased out-of-domain test set (``OOD-Para''), where samples come from ``unseen domains'' and ``unseen models,'' augmenting the detector with discourse motifs yields significant improvements over the baselines.


\begin{table}[!ht]
\centering
\resizebox{0.95\columnwidth}{!}{
\begin{tabular}{l|c|ccc}
\toprule
\multirow{2}{*}{\textbf{Models}} & \multicolumn{1}{c|}{\textbf{HC3}} & \multicolumn{3}{c}{\textbf{DeepfakeTextDetect}} \\
& \textbf{Test} & \textbf{Test} & \textbf{OOD} & \textbf{OOD-Para} \\
\midrule
RF (All Motifs)          & 0.55          & 0.57          & 0.48          & 0.60 \\
RF (Motifs)         & 0.55          & 0.58          & 0.49          & 0.61 \\
\midrule
GAT                      & 0.67          & 0.68          & 0.66          & 0.53 \\
GAT+Motifs        & 0.73          & 0.72          & 0.67          & 0.54 \\
\midrule
LF                       & 0.97          & 0.90          & 0.74          & 0.60 \\
LF+Motifs         & \textbf{0.98} & \textbf{0.93} & \textbf{0.82} & \textbf{0.62} \\
\bottomrule
\end{tabular}}
\caption{Overall detection F1 scores of models on the benchmark datasets. ``RF (All Motifs)'' denotes Random Forest models with all of the found motifs (3,055 of them) considered as inputs. ``RF (Motifs)'' indicates Random Forest models where only the MF-IDF-selected motifs (207 of them) are taken as inputs.\vspace{-1mm}}
\label{tab:overall_f1}
\end{table}

\paragraph{\textsc{TenPageStories}.}\label{subsubsec:creative_writing}
In a more open-ended exploration, we deploy the trained LF model augmented with discourse motifs to analyze our \textsc{TenPageStories} dataset.

\begin{table}[!ht]
\centering
\resizebox{0.9\columnwidth}{!}{
\begin{tabular}{l|ccc|ccc}
\toprule
\multirow{2}{*}{\textbf{Models}} & \multicolumn{3}{c|}{\textbf{FIG}} & \multicolumn{3}{c}{\textbf{Constrained FIG}} \\
                         & \textbf{1} & \textbf{3} & \textbf{5} & \textbf{1} & \textbf{3} & \textbf{5} \\
\midrule
LF               & 0.30 & 0.40 & 0.45 & 0.59 & 0.50 & 0.55 \\
LF+Motifs & \textbf{0.43} & \textbf{0.65} & \textbf{0.71} & \textbf{0.71} & \textbf{0.69} & \textbf{0.70} \\
\bottomrule
\end{tabular}}
\caption{Overall detection F1 scores on \textsc{TenPageStories} where ``FIG'' refers to the ``fill-in-the-gap'' setting. The numbers 1, 3, and 5 indicate the number of paragraphs to be generated (or ``filled in'').}
\label{tab:overall_f1_ten_page_stories}
\end{table}

First, we check how the different generation settings affect the detection performance (Table \ref{tab:overall_f1_ten_page_stories}). As expected the detection model performs better with longer generated texts (i.e., 1, 3, and 5 paragraphs). Also, when the first and last sentences of the human-written texts are additionally provided in the prompt (i.e., ``constrained FIG''), the corresponding generations are easier to detect. We note that the models' f1 scores are low as they produce a lot of false negatives, i.e., predicting actual human-written texts to be machine-generated. However, we can observe that the addition of a small discourse vector significantly improves the performance.



\begin{figure}[!ht]
\centering
{\includegraphics[width=\columnwidth,trim={0cm 0cm 0cm 0.55cm},clip]{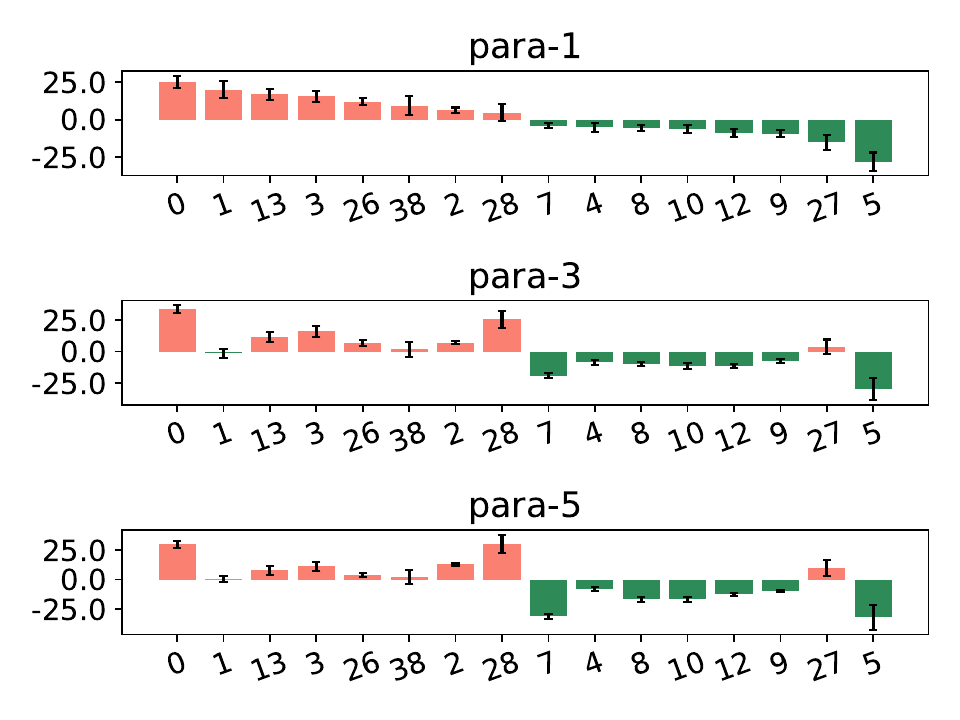}\vspace{-3mm}}
\caption{Difference distribution of motifs for \textsc{TenPageStories} under the \textit{fill-in-the-gap} settings. The x-axis represents unique indices of single-triads while the y-axis shows the difference in motif frequency (scaled by 1e-3) of machine-generated and human-written texts for each motif. Discourse motifs indexed at 0 (Elaboration), 5 (Joint), 7 (Joint), and 28 (Temporal) seem to be useful in distinguishing the two groups.\vspace{-3mm}}
\label{fig:diff_dist_ten_page_stories}
\end{figure}

Similarly, Figure \ref{fig:diff_dist_ten_page_stories} elucidates the distinct motif distributions within \textsc{TenPageStories} across various settings.  It shows that as the generation length increases, motifs indexed at 0 (Elaboration) and 28 (Temporal) become increasingly prevalent in machine-generated texts. Conversely, motifs indexed at 7 (Joint) and 5 (Joint) appear more frequently in texts authored by humans. This may imply that human authors tend to construct narratives with a higher prevalence of Joint relations, indicative of more evenly branching structures, in contrast to the patterns found in texts generated by LLMs.

\subsection{Structural Variations after Paraphrasing}\label{subsec:upper_vs_lower}

\begin{figure}[!ht]
\centering
{\includegraphics[width=\columnwidth,trim={0cm 0cm 0cm 0cm},clip]{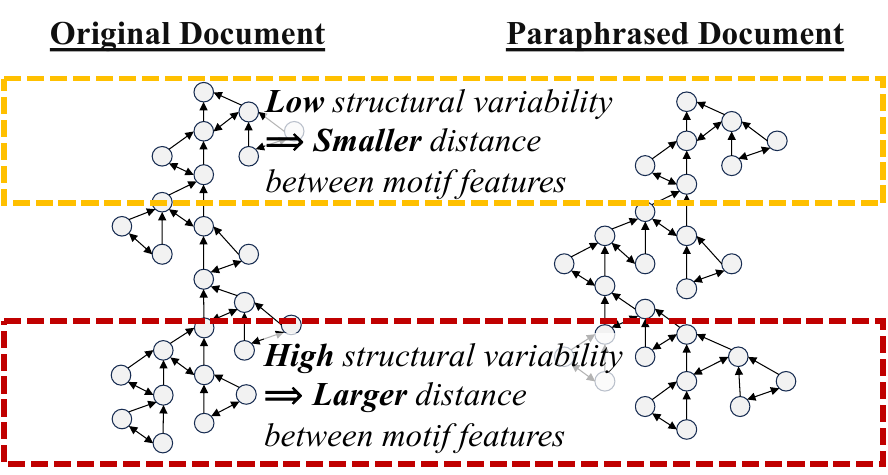}\vspace{-3mm}}
\caption{We posit that sentence-level paraphrasing typically leads to greater variability in lower segments of discourse structures compared to higher segments near the root node.\vspace{-3mm}}
\label{fig:orig_vs_para}
\end{figure}

In this experiment, we look at the discourse structures of original and paraphrased documents regardless of their authorship. We aim to validate our hypothesis that sentence-level paraphrasing induces greater variability in the lower segments of discourse graphs, as these segments correspond to the individual EDUs and their aggregation into sentences. Given the assumption that the overall higher-level discourse structure remains relatively intact, we anticipate that the upper segments will display less structural variability (c.f. Figure \ref{fig:orig_vs_para}).
To this end, we calculate the discourse motif features (\S\ref{subsubsec:motif_diff_dist}) for both upper and lower segments of the graphs corresponding to each pair of original and paraphrased documents. Here, the upper segment denotes the neighborhood graph of the root node within an edge distance of 3. Similarly, the lower segment encompasses the union of neighborhood graphs of individual EDU nodes within one edge distance. We average the motif features over the original and paraphrased documents and compute the absolute distance between the two groups.

\begin{figure}[!ht]
\centering
{\includegraphics[width=\columnwidth,trim={0cm 0cm 0cm 0cm},clip]{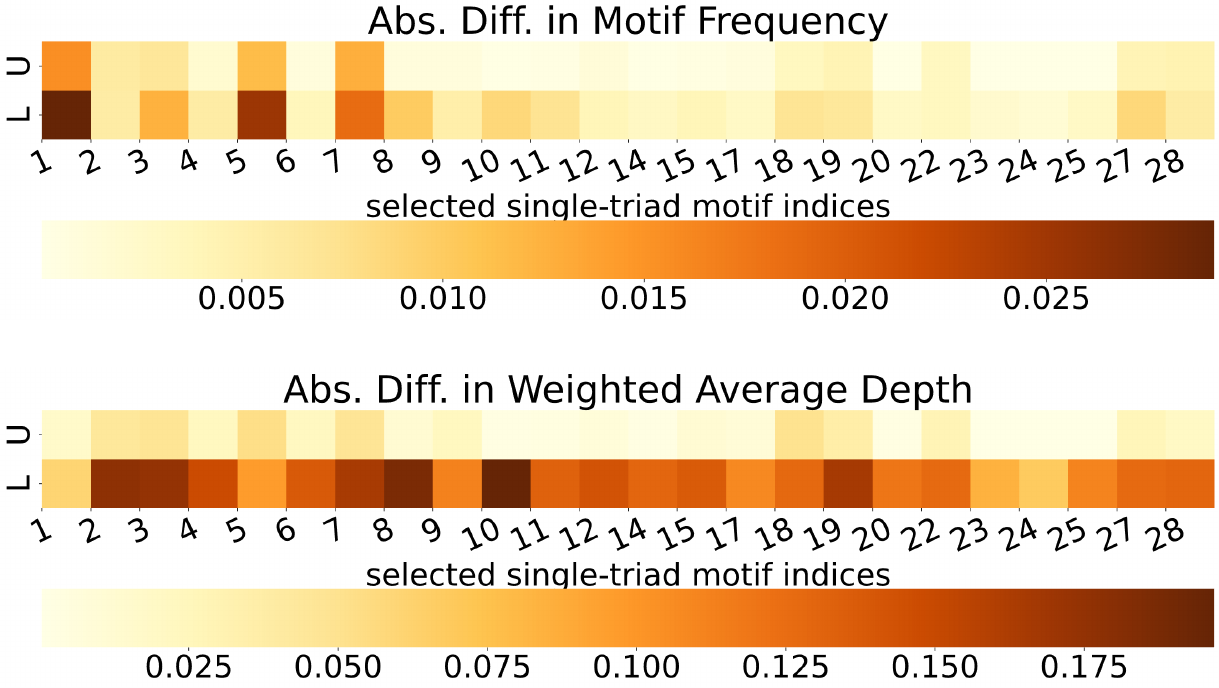}\vspace{-1mm}}
\caption{Two heatmaps illustrating the absolute differences of normalized motif counts (above) and weighted average depth (below) between motifs present in discourse graphs of original and paraphrased texts. A graph is further divided into two separate subgraphs (i.e., ``upper (U)'' and ``lower (L)'') depending on the distance from the root node.\vspace{-3mm}}
\label{fig:upper_vs_lower_discourse}
\end{figure}

Figure \ref{fig:upper_vs_lower_discourse} depicts the results, delineated by motif frequency and weighted average depth. Notably, both features indicate that the lower segments of graphs display a greater absolute distance compared to the upper portions, thereby supporting our hypothesis.

\subsection{Formality Scores and Hyperedges}\label{subsec:formality_scorer}
Concerning text styles, we note that the most noticeable disparity among the various text domains appears to lie in the formality dimension. Thus, we computed the formality score for every text generated by humans and machines (c.f. \S\ref{subsec:baseline}). Also, for every corresponding document graph, we count the (normalized) frequency of hyperedges for all motifs present in the graph. This is because the hyperedge relations represent higher-level abstractions of text and are thus linked to the discourse hierarchy. 
From there, we compute the Pearson correlation coefficient between the scores and frequencies.
As shown in Table \ref{tab:corr_formality_hyperedge}, the human-written texts over all domains exhibit weak to moderate correlation with the formality style while machine-generated ones seem to show no correlation. This hints that when humans write more structured texts in terms of discourse, the contents are more formal.

\begin{table}[!ht]
    \centering
    \resizebox{0.7\columnwidth}{!}{
    \begin{tabular}{c|c}
        \toprule
        \textbf{Machine-Generated} & \textbf{Human-Written} \\
        \midrule
        $0.08$ & $0.39$ \\
        \bottomrule
    \end{tabular}}
    \caption{Pearson's $R$ between formality scores and frequency of hyperedges.\vspace{-3mm}}
    \label{tab:corr_formality_hyperedge}
\end{table}

\subsection{Other Experiments}
We conduct preliminary analyses on discourse motifs and investigate the correlation between document graph shapes across different domains in Appendices \ref{sec:app:motif_analyses} and \ref{sec:app:branch_vs_chain}, respectively.
\section{Conclusion}\label{sec:conclusion}
As texts generated by LLMs become increasingly challenging to differentiate from those authored by humans, our approach involves identifying discernible, unique properties inherent in human-crafted texts. We posit that these distinctive signatures manifest at levels beyond mere surface structure and thus opt for looking into their hierarchical discourse structures. Viewing the structures as recursive hypergraphs and conducting motif analyses on them, we find that these motifs are useful in the authorship detection task and can highlight the subtle structural differences in the two author groups depending on the domains of texts and their corresponding nature. Future plans include extending this approach to long documents by merging multiple document graphs and incorporating topological information beyond discourse.
\section{Limitations}\label{sec:limitation}
In our study, the validity of our experiments and subsequent findings relies on the parsed discourse trees generated by an existing parser. It is important to acknowledge the potential for using alternative discourse frameworks like Segmented Discourse Representation Theory \citep{lascarides2007segmented} though finding or training a robust parser is a challenge.
Despite employing a state-of-the-art model for RST parsing, it is crucial to recognize that parsed results may still be suboptimal due to the inherent difficulty of hierarchical discourse parsing. This challenge is exacerbated by the limited scale of existing datasets used for model training and the inherent ambiguities present in discourse relations. In this aspect, an important future direction would be to build a more robust discourse parser using current LLMs.

In our experimental setup, motif analyses were carried out using single-, double-, and triple-triads. It is worth considering that longer and potentially significant patterns may exist beyond these sizes. However, expanding the motif sizes within our current computational approach is not feasible. Nevertheless, there is potential to capture the distributions of longer patterns through approximation techniques, such as subsampling or leveraging deep learning methodologies.

In the experiment investigating the correlation between formality scores and the number of hyperedges (Section \ref{subsec:formality_scorer}), it is important to acknowledge a limitation: the formality scorer had been trained on short sentence-level inputs and may not be adequately suited for assessing the formality of longer documents spanning multiple paragraphs.

Furthermore, our study highlights the need for more comprehensive analysis within the \textsc{TenPageStories} dataset. Exploring the correlation between discourse structures and specific linguistic features tailored to creative writing, such as tagged events between named entities, presents an intriguing avenue for future research.

\section{Ethical Statement}
While our research aims to distinguish linguistic features in human-written and machine-generated texts to improve authorship detection, it is important to acknowledge the potential for these features to be utilized in developing LLMs capable of generating texts that more closely resemble human-authored ones. To address this concern, we advocate for the implementation of transparent reporting practices, ethical review processes, responsible use policies, and collaboration among stakeholders to ensure the ethical development and deployment of LLMs.

Throughout the paper, we have referenced datasets and tools utilized in our experiments, ensuring they originate from open-source domains and do not pose any conflicts with the public release or usage of this paper.
Our results are also consistent with the licensing terms of the open-source domain from which the datasets and tools were sourced. We also note that our constructed dataset, \textsc{TenPageStories}, stems from fictional ebooks available in the public domain and contains no information that names or uniquely identifies individual real people, nor does it include any offensive content.

We acknowledge the use of Grammarly and ChatGPT 3.5 for correcting any less fluent sentences but not for generating new content.

\section*{Acknowledgements}
This research was primarily funded by a generous research gift from Grammarly. Kwang Hee Lee received funding from the National Research Foundation of Korea (NRF), which is supported by the Korean government (MSIT), under Grant No. RS-2024-00345567. We are grateful to the Minnesota NLP group members for their valuable feedback and constructive comments on our initial draft. We also acknowledge Jonghwan Hyeon and Chae-Gyun Lim from KAIST for their assistance with the design of the figures.

\bibliography{anthology,custom}

\appendix
\onecolumn
\section{Comparison with prior work}\label{sec:related_work}
With the advent of recent LLMs, the need, as well as the challenges in distinguishing machine-generated texts, has risen significantly. Various existing methodologies encompass fine-tuning LLMs \citep{guo2023close,liu2023argugpt,liu2023check,li2023deepfake} with supervised datasets, potentially utilizing techniques like contrastive learning \citep{liu2023coco,bhattacharjee2023conda}, adversarial learning \citep{shi2023red,yang2023chatgpt}, or human-assisted learning \citep{uchendu2023does,dugan-etal-2023-roft}. Additionally, there are zero-shot methods \citep{Beresneva2016ComputerGeneratedTD,vasilatos2023howkgpt} that leverage statistical features of texts or intermediate values within LLMs; watermarking techniques \citep{zhao-etal-2022-distillation,pmlr-v202-kirchenbauer23a} that introduce ``green tokens'' or embed ``secret keys'' as vectors to generated outputs; and prompt-based approaches that make use of (another) LLM as detectors \citep{{Zellers2019DefendingAN,Koike2023OUTFOXLE,Yu2023GPTPT,bhattacharjee2023fighting}}. A comprehensive overview of these approaches can be found in \citet{wu2023survey}.

While the relevant recent studies on the detection task primarily adopt surface-level features, such as distributions of n-grams and part-of-speech (POS) tags, we approach the problem from another direction: using hierarchical discourse features within texts. This direction is reminiscent of the work by \citet{corston-oliver-etal-2001-machine}, who explored a comparable path in the pre-LLM era. Their study investigated the efficacy of the branching features observed in syntactic parse trees to determine the origin of texts, particularly whether they were generated by a machine translation model. Our work considers discourse-level features spanning multiple sentences and paragraphs (as opposed to sentence-level POS features).

The existing challenges in the field encompass several facets. First, there are pronounced out-of-distribution (OOD) challenges \citep{li2023deepfake,antoun-etal-2023-towards}, wherein the detectors struggle when confronted with texts that fall outside the learned distribution. Second, the detectors are prone to potential attacks \citep{shi2023red,he2024mgtbench}, such as paraphrasing attacks \citep{sadasivan2023aigenerated,krishna2023paraphrasing}, where machine-generated texts undergo further paraphrasing to alter the distribution of lexical and syntactic features. This dynamic renders detectors reliant on surface-level features and watermarking technology ineffective. 
Finally, the inherent ambiguities present in the two groups of texts have become \textit{more nuanced} over time, making it progressively challenging to distinguish them. In this paper, we touch upon these three challenges by showing that the addition of discourse features (i) improves the detection of OOD samples and (ii) trains a more robust classifier against paraphrased attacks. Furthermore, our analyses, utilizing discourse network motifs, shed light on the nuanced distinctions within the hierarchical structures of the two text categories.


\section{Recursive hypergraphs to their standard traditional graph form}\label{sec:app:unions}
\begin{figure*}[!h]
  \includegraphics[width=\textwidth]{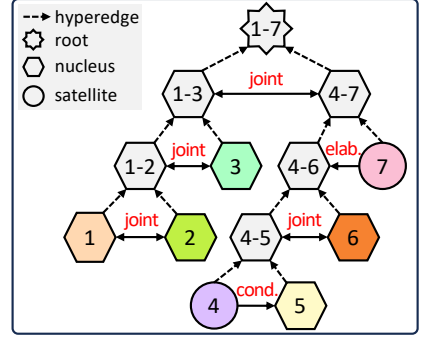}
  \caption{Steve Job's motivational quote and its corresponding RST tree.}
  \label{fig:rst_tree_example}
\end{figure*}

In this section, we introduce how recursive hypergraph representations of RST trees can be transformed into standard traditional graph forms.
Suppose an RST tree consists of $m$ EDUs $\{\textrm{EDU}_1, \cdots, \textrm{EDU}_m  \}$ with their relations such as $(\textrm{EDU}_1, \textrm{EDU}_2)$ and $((\textrm{EDU}_1, \textrm{EDU}_2), \textrm{EDU}_3)$ (see EDU1, EDU2, EDU3 in Figure \ref{fig:rst_tree_example} for example).
Note that in this section, we ignore edge labels for better readability.
A recursive hypergraph representation of the given RST in Figure \ref{fig:rst_tree_example} is a tuple $(V, E)$ where 
 a set of vertices $V=\{\textrm{EDU}_1, \cdots, \textrm{EDU}_7  \}$ and
  a set of edges $E=\{e_{1-2}=(\textrm{EDU}_1, \textrm{EDU}_2), e_{1-3}=(e_{1-2}, \textrm{EDU}_3),  \cdots,  e_{1-7}=(e_{1-3}, e_{4-7})\}$.\footnote{We can consider bi-direction by adding a reverse order of an edge.}
  To facilitate ease of analysis, the hypergraph can be transformed into a standard traditional graph form as follows.
  \newtheorem{definition}{Definition}
  \begin{definition}
  A standard traditional graph form $(V', E')$ of a recursive hypergraph representation $(V, E)$ of an RST tree is defined by $V'=V\cup E$ and $E'=\{(u, w) | (u, w) \in E \textrm{ or } u \in w\}$.
  \end{definition}
 This graph is thought of as an ``extended'' Levi graph of a recursive hypergraph \cite{Joslyn2017UbergraphsAD}.\footnote{Different from the definition of Levi graphs in \cite{Joslyn2017UbergraphsAD}, we allow cycles in the extended Levi graphs.} 
We call it simply the transformed graph of an RST tree if there is no ambiguity.

In the definition of the transformed graph, each relation in an RST tree constitutes a triangle graph. A triangle graph is a graph with three vertices where there is at least one edge between two vertices. For example, a relation $r=(x, y)$ ($x$ and $y$ are either an EDU or an edge) is transformed to the triangle graph, $V'=\{r, x, y\}, E'=\{(x, y), (x, r), (y, r)\}$. Now, we generalize this observation as follows. 

  \newtheorem{theorem}{Theorem}
  \begin{theorem}
  The transformed graph of an RST tree only consists of the union of triangle graphs.
  \end{theorem}

\textbf{Proof sketch by induction}.
Before we prove, assume that every finite RST can be represented by a huge, single relation in a recursive manner. For example, the RST in Figure \ref{fig:rst_tree_example} can be represented by 
\begin{equation}
    (((\textrm{EDU}_1, \textrm{EDU}_2), \textrm{EDU}_3), (\textrm{EDU}_7, ((\textrm{EDU}_4, \textrm{EDU}_5), \textrm{EDU}_6))).
\end{equation}


Consider an RST tree $(\textrm{EDU}_1, \textrm{EDU}_2)$ as an initial case. Then, the transformed graph is a triangle graph.

Suppose $x$ that is an RST tree with $n-1$ EDUs is the union of triangle graphs. Then, an RST tree $(x, \textrm{EDU}_n)$ or $(\textrm{EDU}_n, x)$ is also a graph with the union of triangle graphs by the definition.

More generally, suppose that an RST tree with less than $n$ EDUs forms the union of triangle graphs. 
For an RST tree $z$ containing $n$ EDUs, this RST can be divided into an RST tree $x$ with $n-k$ EDUs and an RST tree $y$ with $k$ EDUs, i.e., $z=(x, y)$. Since both transformed graphs of $x$ and $y$ are unions of triangle graphs by the assumption, and a relation $(x, y)$ can be transformed into the triangle graph as discussed above, the RST $z$ can be also transformed into a new graph with unions of triangle graphs.


\newpage
\section{Calculation of motif features}\label{sec:app:motif_feats_calc_example}
\begin{figure}[!ht]
\centering
{\includegraphics[width=0.9\textwidth]{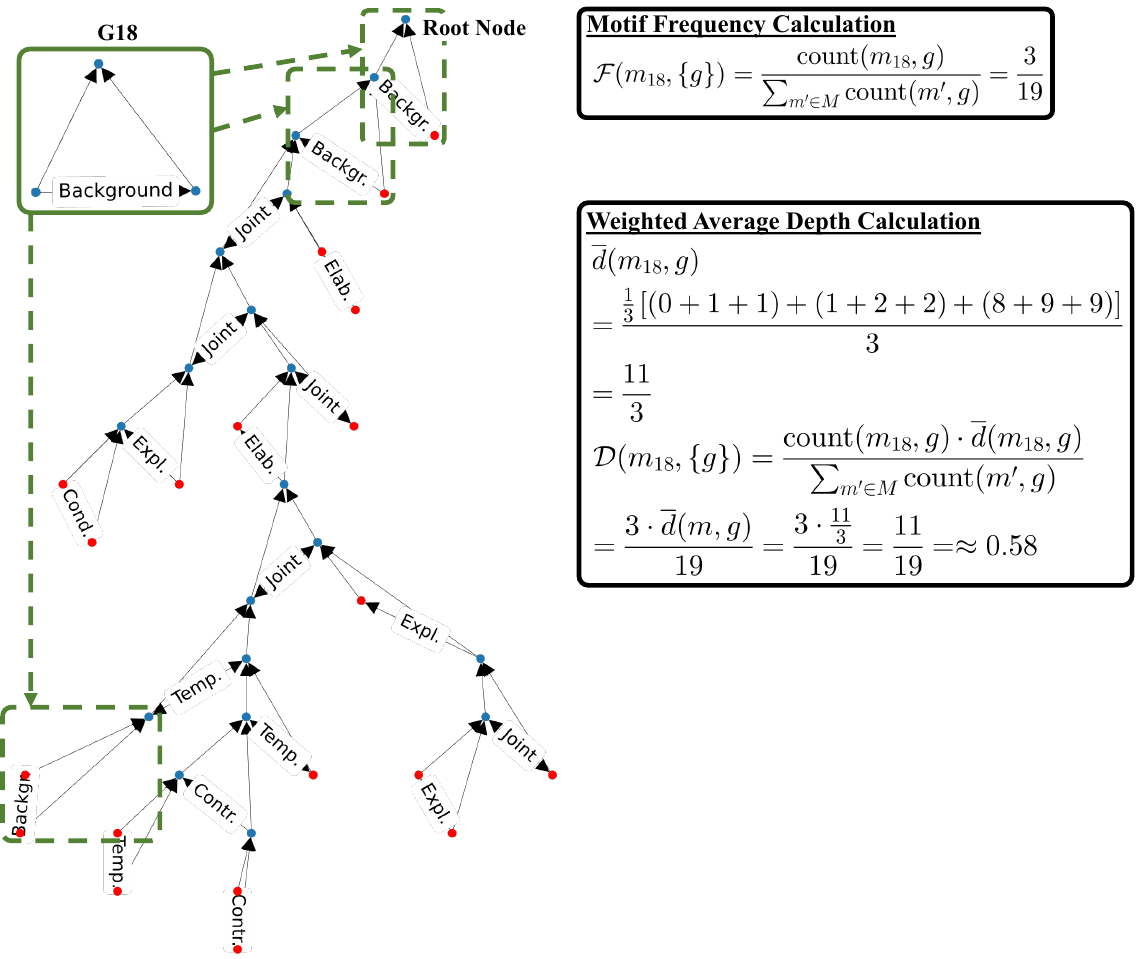}}
\caption{An example of calculating motif frequency and weighted average depth for the single-triad $m_{18}$.\vspace{-5mm}}
\label{fig:feature_calc}
\end{figure}

Figure \ref{fig:feature_calc} illustrates the calculation of the two features, motif frequency and weighted average depth, of a discourse motif with index 18. This motif is one of the selected single-triads that exhibit MF-IDF scores (\S\ref{subsubsec:mfidf}) surpassing at least one standard deviation.

\section{Preliminary analyses on discourse motifs}\label{sec:app:motif_analyses}
We begin our preliminary analyses by examining the relationship between the proposed discourse motifs and influential edges identified by the GAT explainer (\S\ref{sec:app:subsec:motifs_and_inf_edges}). Subsequently, we calculate the difference distributions of motifs across five domains to emphasize their distinctive features (\S\ref{sec:app:subsec:key_motifs}).

\subsection{Correlation between Motifs and Influential Edges for Explanation}\label{sec:app:subsec:motifs_and_inf_edges}
As a way to evaluate the appropriateness of using discourse motifs, we explore how they are related to influential edges in GATs computed by an explanation method. Specifically, we utilize explainability techniques such as ``GNNExplainer'' \citep{ying2019gnnexplainer} and ``GraphFramEx'' \citep{amara2022graphframex} for graph classification, aiming to assess their correlation with our proposed motif features. These methods elucidate the predictions of GNNs by identifying the nodes or edges whose masking has the most significant impact on the predicted outcome. Since our graphs consist of directed and labeled edges, we require an explanation method tailored to edge-based features for GAT. Thus, we employ ``AttentionExplainer'' from GraphFramEx, which computes edge masks using the attention coefficients generated by GATs trained on the binary classification task.

Note that the influential edges refer to edges with high masking values\footnote{The most influential edges whose masking values are larger than $0.99$ are considered in $[0, 1]$.} generated by a GAT explainer as described in (\S\ref{sec:exp_setup}).
We then calculate a correlation between frequencies of selected discourse motifs based on MF-IDF scores and those of influential edges.
To this end, we define two random variables as follows. Let $X$ be the number of single-triads and let $Y$ be the number of influential edges for each graph.
Samples of $X$ and $Y$ are collected from each graph in the datasets (\textsc{HC3-English}, \textsc{DeepfakeTextDetect}), and the Pearson correlation coefficient $r_{XY}$ is calculated. 

\begin{figure}[!ht]
\centering
{\includegraphics[width=0.5\textwidth]{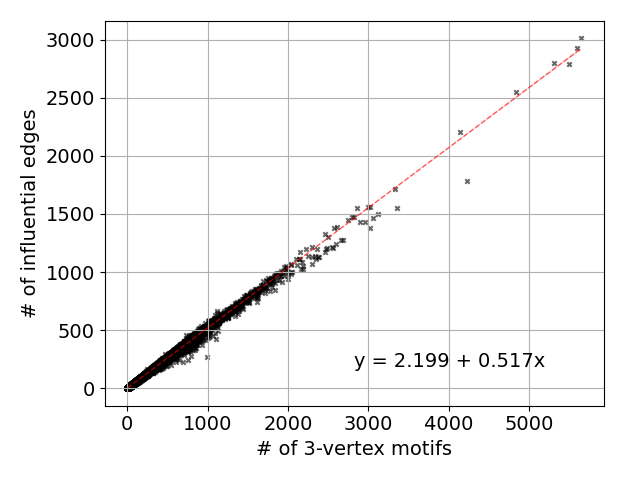}\vspace{-3mm}
}
\caption{Correlation between selected single-triads with high MF-IDF scores and influential edges computed by a GAT explainer on \textsc{DeepfakeTextDetect} dataset.\vspace{-3mm}}
\label{fig:corr_m3}
\end{figure}

\begin{figure}[!ht]
\centering
{\includegraphics[width=0.5\textwidth]{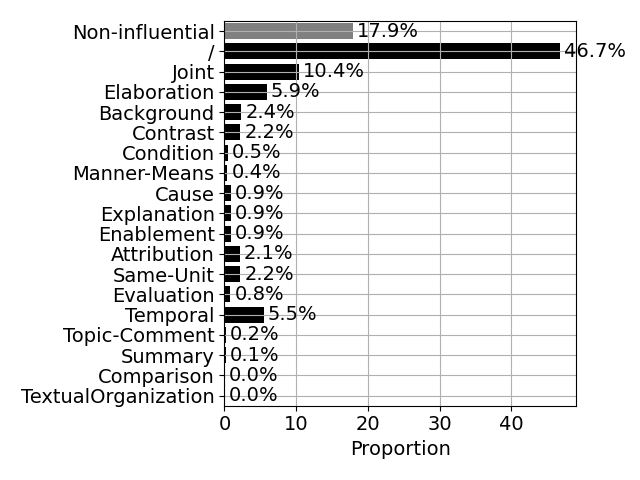}\vspace{-3mm}}
\caption{Proportions of labels of influential edges. Note that ``Non-influential'' refers to edges whose mask values are under the threshold. Note also that the value ``0.0\%'' in this figure does not indicate the zero, but a very small positive number like ``1e-5\%.''\vspace{-3mm}}
\label{fig:proportion_edge_label_m3}
\end{figure}

Figure \ref{fig:corr_m3} shows that frequencies of selected single-triads and influential edges are highly correlated with $r_{XY} \approx 0.99$, indicating that the discourse motifs may be useful features in determining the authorship of texts. 
Figure \ref{fig:proportion_edge_label_m3} shows the proportion of edge labels (i.e., discourse relations) of influential edges. We note that the label ``/'' in the figure denotes the hyperedge relation (``hyp.'').

It is noteworthy that while the motifs are motivated by the discourse framework rooted in linguistics, the notion of influential edges stems from a machine learning-driven, data-centric approach. The strong correlation observed between the two is particularly interesting.

\subsection{Difference Distributions of Motifs across Domains}\label{sec:app:subsec:key_motifs}

\begin{figure*}[!ht]
  \centering
  \includegraphics[width=0.5\textwidth, trim={2cm 0.0cm 0cm 0cm},clip]{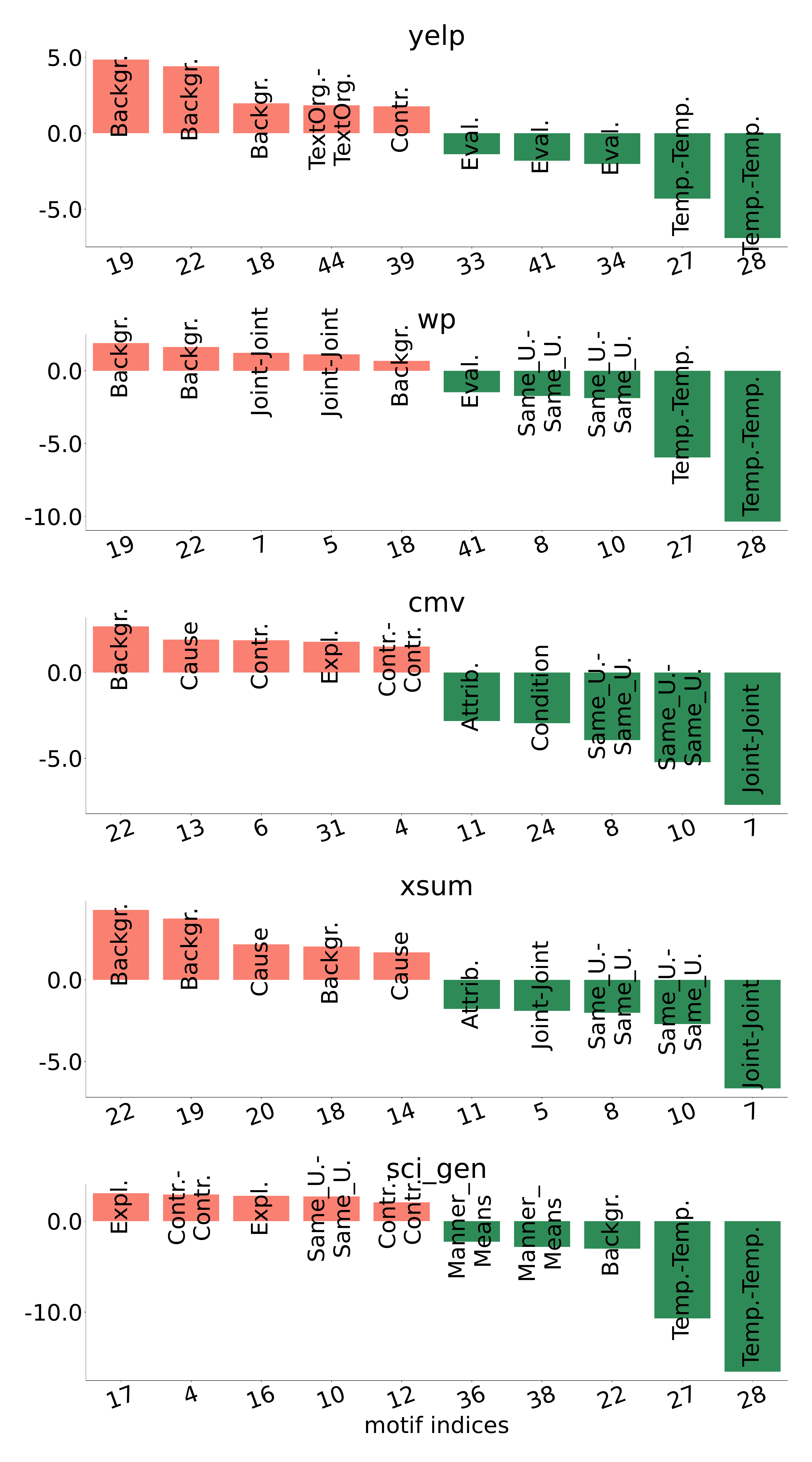}
  \caption{Difference in single-triad motif distribution of machine-generated and human-written texts for five different domains in \textsc{DeepfakeTextDetect} dataset. Y-axis represent the difference in MF-IDF scores scaled by 1e-3. Full-scaled versions including a larger number of motifs and plots for double- and triple-triads can be found in Appendix \ref{sec:app:more_diff_dists}.}
  \label{fig:found_motifs}
\end{figure*}

Figure \ref{fig:found_motifs} shows bar plots for the difference distribution of single-triad motifs across the five domains of texts. We note that large-scaled plots can be found in Figure \ref{fig:large_motifs_dists}.

The mean and standard deviation of differences ($\mathcal{M}(\mathcal{D}_{\text{diff}})$) are is 0.00012 and 0.0015, respectively. Notably, machine-generated texts tend to feature more motifs with Background relations, while human-written texts exhibit an inclination towards Temporal discourse relations in domains such as review writing (\textsc{yelp}) and story generation (\textsc{wp}), and contain more Joint relations in domains that demand formality or deeper logical depths (i.e., argument writing (\textsc{cmv}) and news summarization (\textsc{xsum}). We note that Joint discourse relation signifies cases where the corresponding child trees are of equal importance (i.e., ``nuclei'') and thus branched evenly.
The texts in table description generation (\textsc{sci\_gen}) are interesting in that the human-written descriptions (taken mostly from academic papers in NLP and machine learning) mainly highlight the changes in performance and this is captured as Temporal, Manner, or Means relations. Table \ref{tab:grounded_examples_all} shows some grounded examples of texts for prevalent motifs in each domain.



\section{Constructing ``TenPageStories'' dataset}\label{sec:app:ten_page_stories}
The ``TenPageStories'' dataset consists of stories with approximate lengths of 8,000 tokens or 10 A4 pages. It was constructed using an iterative method of calling the OpenAI's \textsc{gpt-4-1106-preview} model. In total, we conducted 552 request calls, resulting in a total expenditure of 55 USD.

The construction process works by continuously adding the text generated by the model in one call to the prompt of the next call. The iterative method was used for all three generation settings: (1) unconstrained generation, (2) ``fill-in-the-gap'', and (3) constrained ``fill-in-the-gap''. Pseudo-code and examples for the three generation settings are provided below. Note that for the constrained fill-in-the-gap example, three different completion calls are made for an iteration: the first completes paragraph 1, the second generates paragraph $2...n-1$, the third completes the last paragraph $n$, where $n$ is the number of paragraphs to be masked per iteration. The exception to this process is when $n=1$, then one completion call is made to complete the paragraph based on its first and last sentences.

\begin{algorithm}[H]
\caption{Unconstrained Generation}
\begin{algorithmic}[1]
\Function{UnconstrainedGeneration}{\texttt{promptInstruction}}
    \State \texttt{prompt} $\gets$ \texttt{promptInstruction}
    \While{\texttt{length(prompt)} $< 8000$ tokens}
        \State \texttt{generatedText} $\gets$ \textit{generate text using LLM based on the prompt}
        \State \texttt{prompt} $\gets$ \texttt{prompt} + \texttt{generatedText}
    \EndWhile
    \State \textbf{return} \texttt{prompt}
\EndFunction
\end{algorithmic}
\end{algorithm}
\vspace{-\intextsep}
\begin{algorithm}[H]
\caption{``Fill-in-the-gap''}
\begin{algorithmic}[1]
\Function{FillInTheGap}{\texttt{promptInstruction, numMasked}}
    \State \texttt{prompt} $\gets$ \texttt{promptInstruction + firstParagraph}
    \For{each group of \texttt{numMasked+1} paragraphs from the second paragraph onwards}
        \State \texttt{generatedText} $\gets$ \textit{generate text using the LLM to fill the first \texttt{numMasked} paragraphs in the group}
        \State \texttt{prompt} $\gets$ \texttt{prompt + generatedText + lastParagraphInTheGroup}
    \EndFor
    \State \textbf{return} \texttt{prompt}
\EndFunction
\end{algorithmic}
\end{algorithm}
\vspace{-\intextsep}
\begin{algorithm}[H]
\caption{Constrained ``fill-in-the-gap''}
\begin{algorithmic}[1]
\Function{ConstrainedFill}{\texttt{promptInstruction, numMasked}}
    \State \texttt{prompt} $\gets$ \texttt{promptInstruction}
    \For{each group of \texttt{numMasked} paragraphs}
        \State \texttt{firstParagraph} $\gets$ \textit{generate text using the LLM to fill the first paragraph in the group based on its first sentence}
        \State \texttt{prompt} $\gets$ \texttt{prompt + firstParagraph}
        \State \texttt{middleParagraphs} $\gets$ \textit{generate text using the LLM to fill all paragraphs in the group besides the first and last}
        \State \texttt{prompt} $\gets$ \texttt{prompt + middleParagraphs}
        \State \texttt{lastParagraph} $\gets$ \textit{generate text using the LLM to fill the last paragraph based on its last sentence}
        \State \texttt{prompt} $\gets$ \texttt{prompt + lastParagraph}
    \EndFor
    \State \textbf{return} \texttt{prompt}
\EndFunction
\end{algorithmic}
\end{algorithm}

\newpage
\begin{enumerate}[noitemsep]
    \item \textbf{Free Generation Example Prompt}: 
    \begin{tcolorbox}[colback=white, arc=4pt]
    Take a look at the story and generate text according to the instructions in brackets []. Only return the generated parts. Here is the story:
    \\\textit{Text from previous generation}
    \\\textit{Text from previous generation}
    \\\textit{Text from previous generation}
    \\$[\text{Generate text here}]$
    \end{tcolorbox}
    \item \textbf{``Fill-in-the-gap'' Example Prompt (Fill 1 paragraphs)}: 
    \begin{tcolorbox}[colback=white, arc=4pt]
    Take a look at the story and generate text according to the instructions in brackets []. Only return the generated parts. Here is the story:
    \\\textit{Original paragraph text}
    \\\textit{Text from previous generation}
    \\\textit{Original paragraph text}
    \\$[\text{Generate paragraph of approximately 94 tokens here}]$
    \\\textit{Original paragraph text}
    \end{tcolorbox}

    \item \textbf{Constrained ``Fill-in-the-gap'' Example Prompt (Fill 3 paragraphs)}: 
    \begin{tcolorbox}[colback=white, arc=4pt]
    Take a look at the story and generate text according to the instructions in brackets []. Only return the generated parts. Here is the story:
    \\\textit{Text from previous generation}
    \\\textit{Text from previous generation}
    \\\textit{Text from previous generation}
    \\\textit{Original first sentence of paragraph 1}
    \\$[\text{Complete preceding paragraph using approximately 67 tokens}]$
    \\$[\text{Generate paragraph of approximately 129 tokens here}]$
    \\$[\text{Complete following paragraph using approximately 87 tokens}]$
    \\\textit{Original last sentence of paragraph 3}
    \end{tcolorbox}
\end{enumerate}


\section{Details on experimental setup}\label{sec:app:exp_details}
Our experiments were conducted on a workstation equipped with 4 NVIDIA RTX A5500 GPUs, an AMD Ryzen Threadripper PRO 5975WX 32-core CPU, and 1TB of RAM. When running experiments, we primarily adhered to the default hyperparameter configurations of the baseline models and model optimizer, with precise details available in the accompanying code repository: \url{[TBA]}.

We report that RF, GAT, and LF models underwent training for 1 epochs ($\approx10$ minutes), 10 epochs ($\approx7.5$ hours), and 4 epochs ($\approx14.5$ hours), respectively, with the best-performing checkpoint chosen based on validation dataset performance.

We also report that parsing RST structures and extracting discourse features across the entire dataset required approximately 9.5 hours, utilizing the multiprocessing capabilities of CPU cores to their fullest extent. We note that these features only need to be computed once offline.


\newpage
\section{More examples of grounded motifs in texts}\label{sec:app:more_grounded_motifs}
Table \ref{tab:grounded_examples_all} shows more examples of texts featuring discourse motifs with high MF-IDF scores.

\begin{table*}[!h]
  \includegraphics[width=\textwidth]{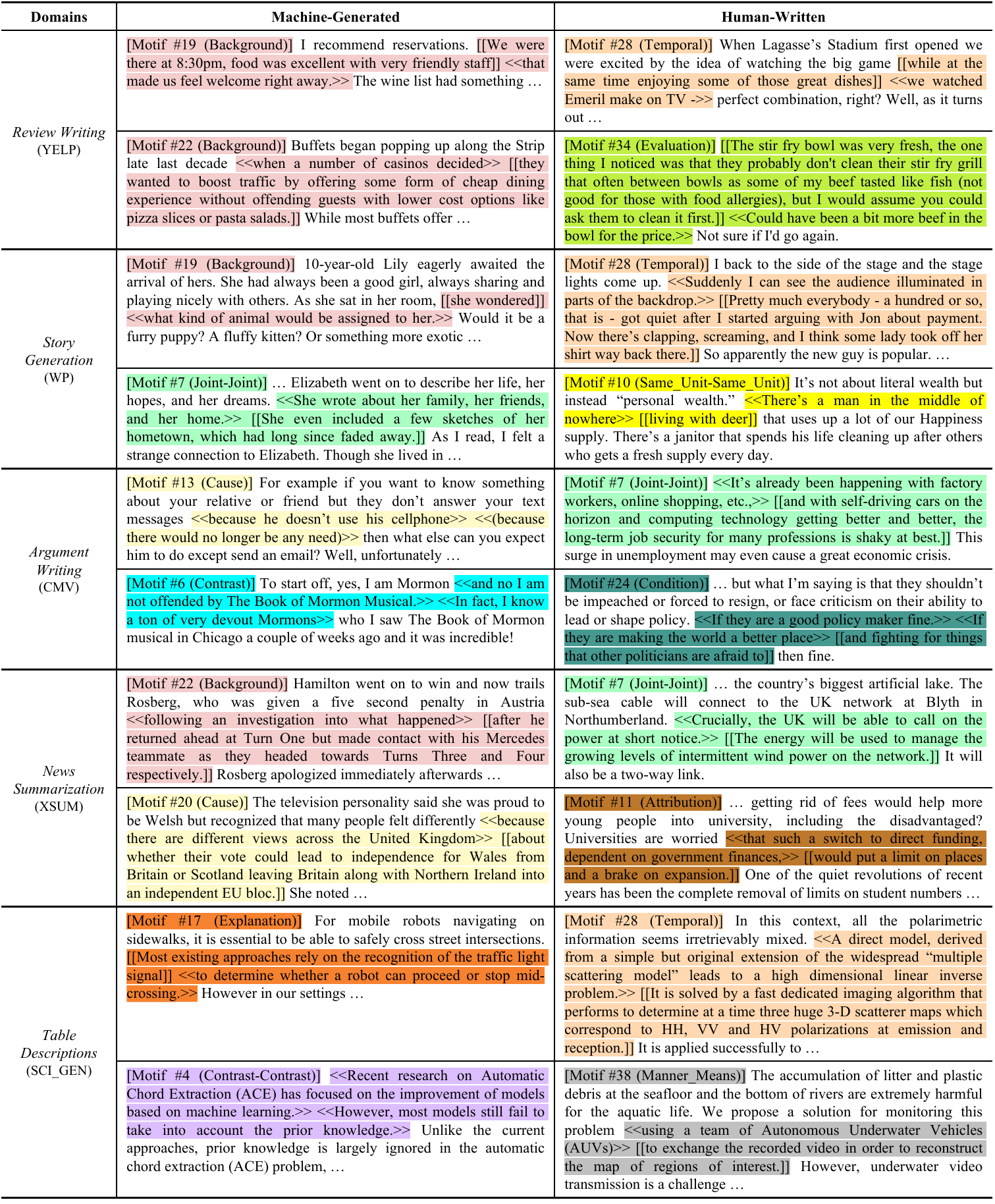}
  \caption{Grounded examples of texts for some prevalent motifs. Exact shapes of the motifs can be found in Fig. \ref{fig:all_motifs_three}.}
  \label{tab:grounded_examples_all}
\end{table*}

\newpage
\section{Additional experiment: branched vs. chain-like structures}\label{sec:app:branch_vs_chain}
In this experiment, we differentiate texts by domain and assume that the formality of texts is consistent within the domain. We can then look at the varying levels of graph structures per domain. As illustrated in the Introduction (\S\ref{sec:introduction}), we assume that the linguistic structure of discourse can change depending on factors such as formality, spontaneity, and depth of reasoning in the texts. 

\begin{figure}[ht!]
  \centering
  \includegraphics[width=0.42\textwidth]{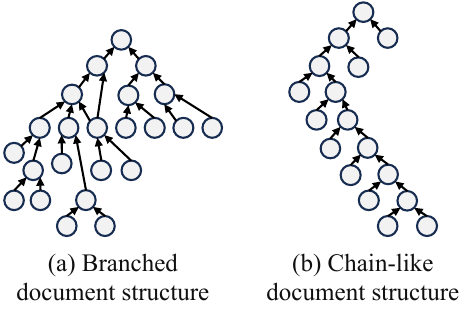}
  \caption{Illustration of document graph structures with different discourse hierarchies. For simplicity, the horizontal edge linking two child nodes is omitted.}
  \label{fig:branch_vs_chain}
\end{figure}

One possible dimension of characterizing this difference in structures is by checking whether the structures are evenly branched or follow a more sequential pattern as illustrated in Figure \ref{fig:branch_vs_chain}.

To quantify this notion, we calculate the average shortest path length (ASPL) of a document graph. For linear or chain-like graphs, the ASPL tends to be relatively short, as nodes are connected in a linear fashion. In contrast, more spread-out graphs will likely have a longer ASPL due to increased distances between nodes. Another closely related metric is the diameter of a graph which is the longest shortest path between any two nodes in the graph. However, as human-written and machine-generated texts are not necessarily paired and could vary a lot in length, we opt for ASPL.

\begin{figure}[!h]
\centering
{\includegraphics[width=0.55\textwidth, trim={0.3cm 0.5cm 0.3cm 0.2cm},clip]{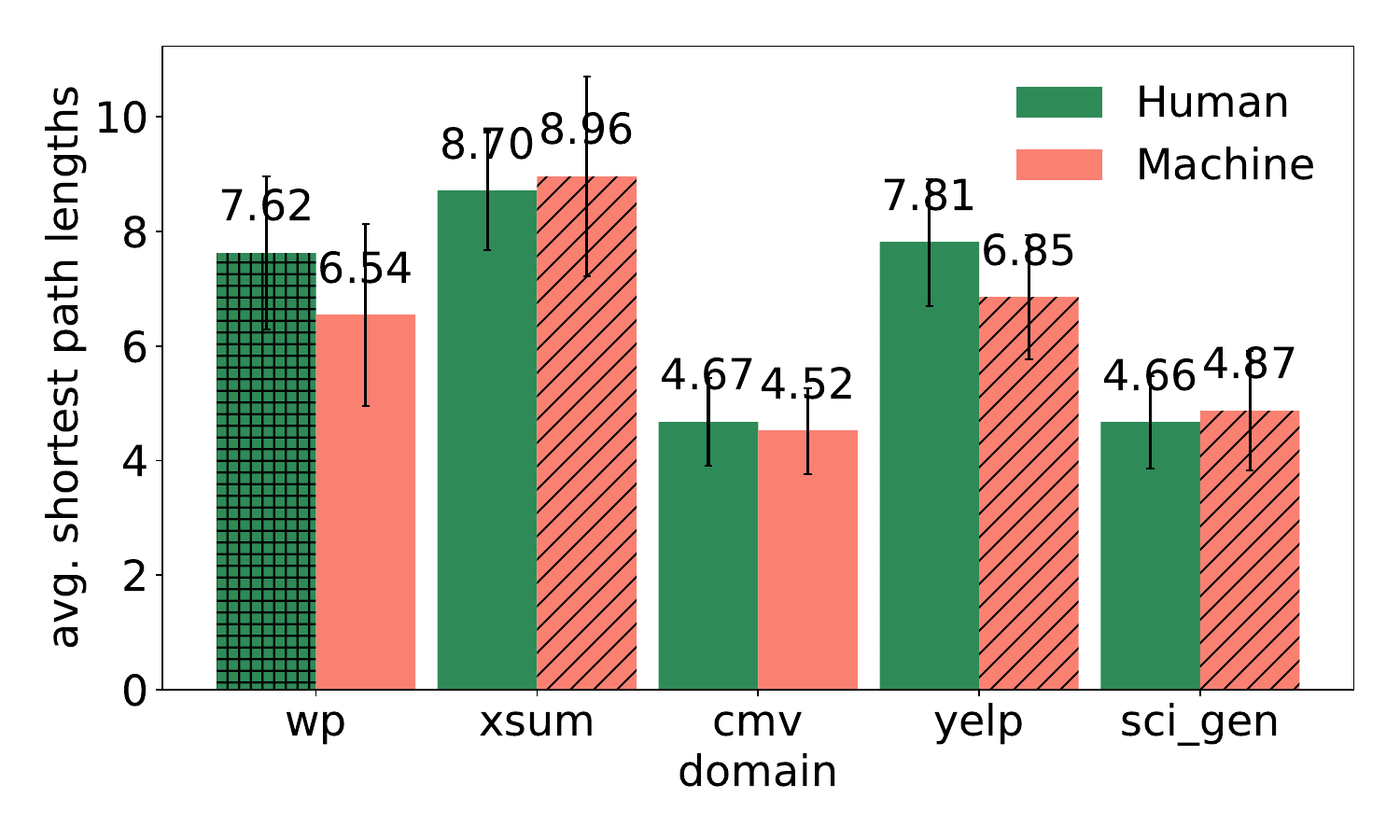}}
\caption{Average shortest path lengths per domain for document graphs. We }
\label{fig:shortest_path_len}
\end{figure}

We observe that the news summarization task (\textsc{xsum}) yields texts with the longest ASPL across both groups.
Human-generated texts exhibit longer ASPL in the \textsc{wp}, \textsc{cmv}, and \textsc{yelp} domains. Notably, within these three domains, Joint discourse relations emerge as the most prevalent relation on the human-authored side, affirming our hypothesis regarding the ``equal branching'' characteristic of this relation (c.f., Fig. \ref{fig:large_motifs_dists}).


\newpage
\section{Difference distributions computed by motif frequency vs. MF-IDF scores}

\begin{figure*}[!ht]
\centering
\begin{tabular}{c}
   {\includegraphics[width=0.9\textwidth]{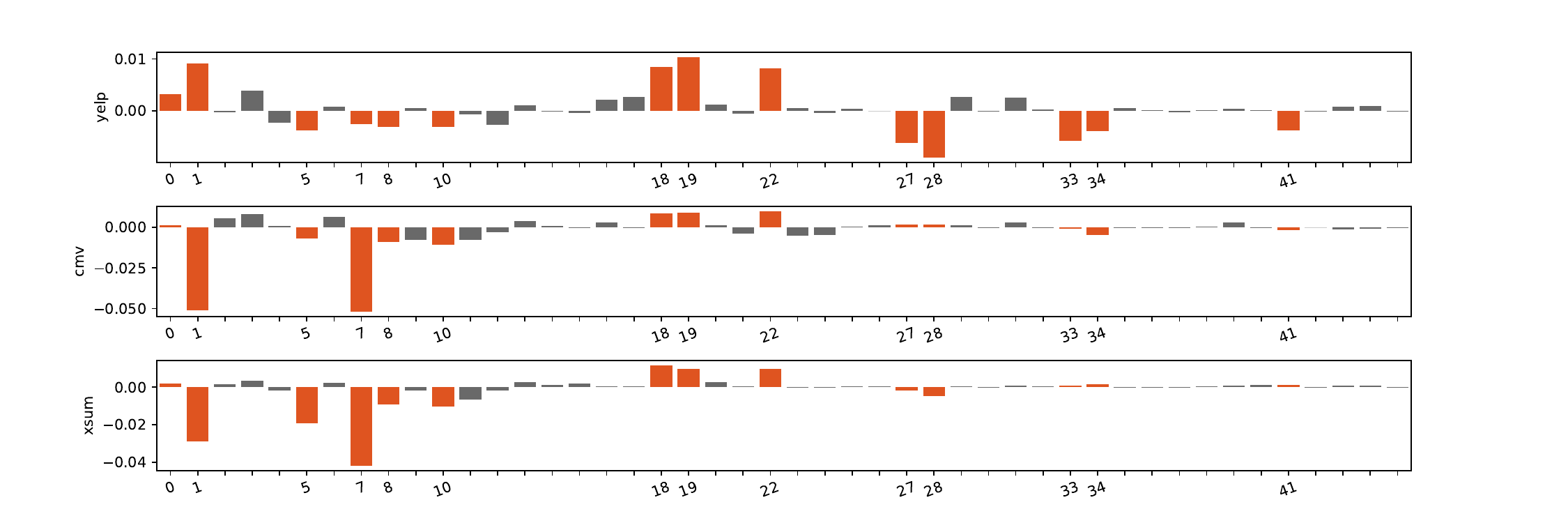}}  \\
     (a) Difference distribution of motifs computed by using their motif frequencies.\\
     \includegraphics[width=0.9\textwidth]{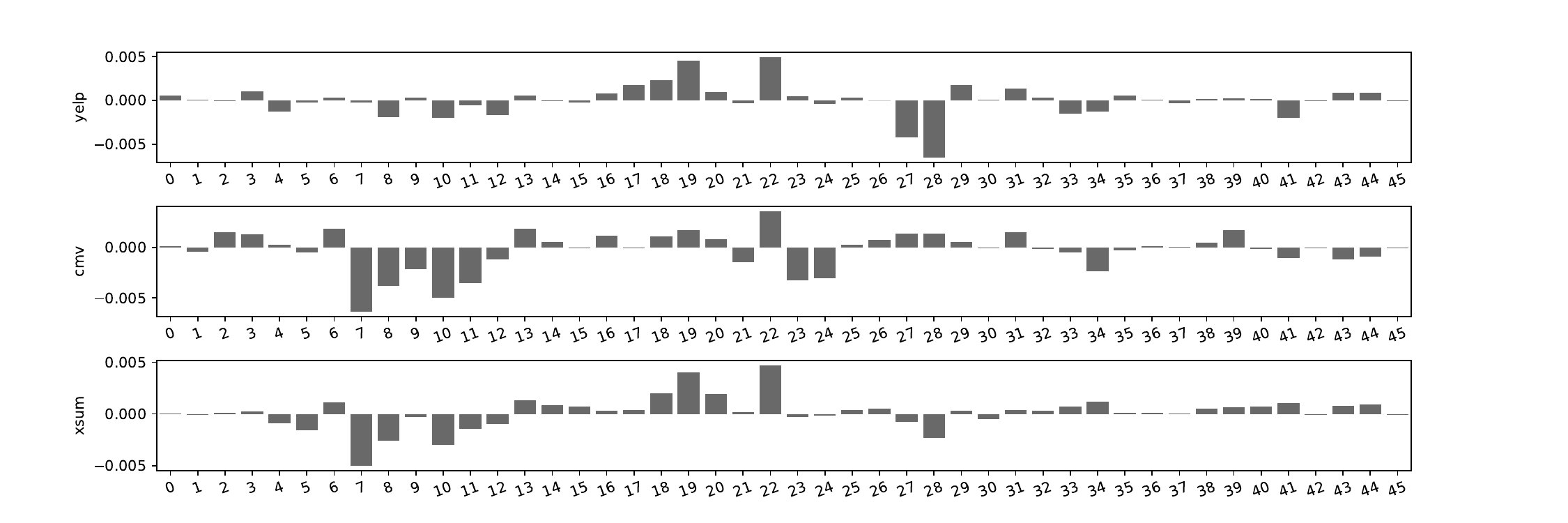}  \\
     (b) Difference distribution of motifs computed by using their MF-IDF scores.
\end{tabular}
\caption{Two ways of computing the difference distributions.}
\label{fig:difference_dists_two_ways}
\end{figure*}

Figure \ref{fig:difference_dists_two_ways} presents two bar plots, illustrating (a) motif frequency and (b) MF-IDF scores of motifs across three domains. Notably, due to the IDF scaling, the latter plot exhibits a slightly less skewed pattern compared to the former plot.

\newpage
\section{Large-scale difference distributions}\label{sec:app:more_diff_dists}
\begin{figure*}[!ht]
\centering
\begin{tabular}{c}
   {\includegraphics[width=0.9\textwidth]{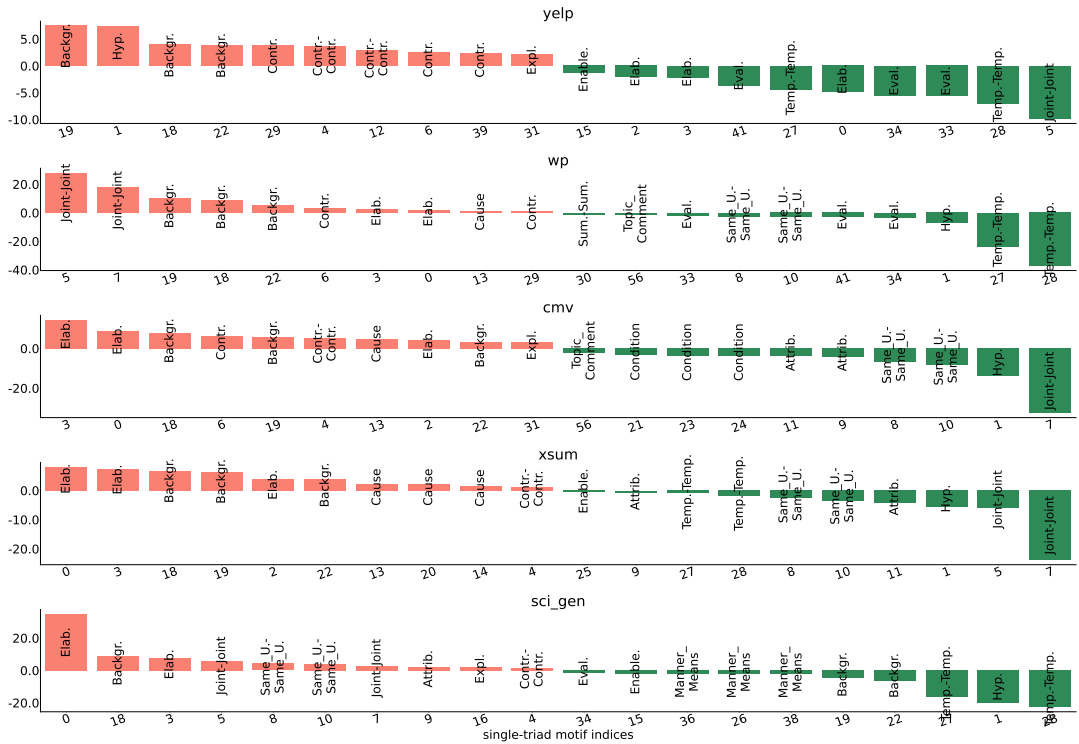}}  \\
     (a) Single-triads.\\
     \includegraphics[width=0.9\textwidth]{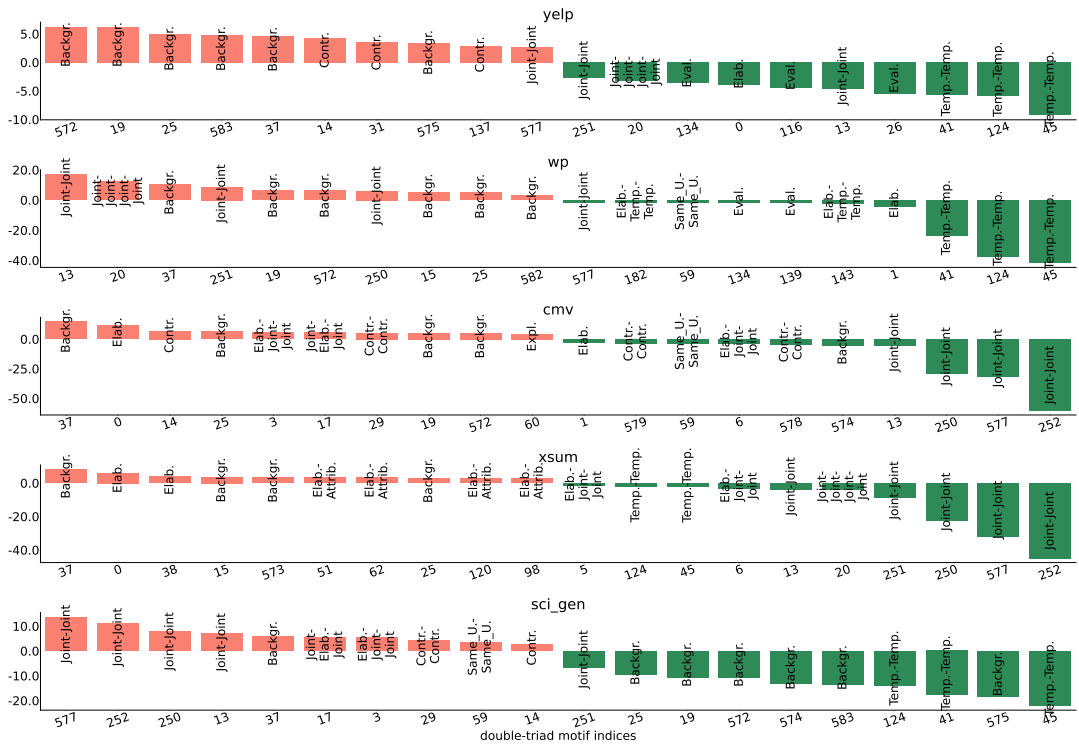}  \\
     (b) Double-triads.
\end{tabular}
\caption{MF-IDF difference distributions of single- and double-triad motifs. To improve readability, the hyperedge relations have been excluded.}
\label{fig:large_motifs_dists}
\end{figure*}

Figure \ref{fig:large_motifs_dists} displays two bar plots for single- and double-motifs, depicting MF-IDF difference distributions. Triple-triads are excluded due to label clutter.

\section{Graphical examples of identified motifs}\label{sec:app:motif_graphs}
\begin{figure*}[!ht]
\centering
{\includegraphics[width=0.43\textwidth]{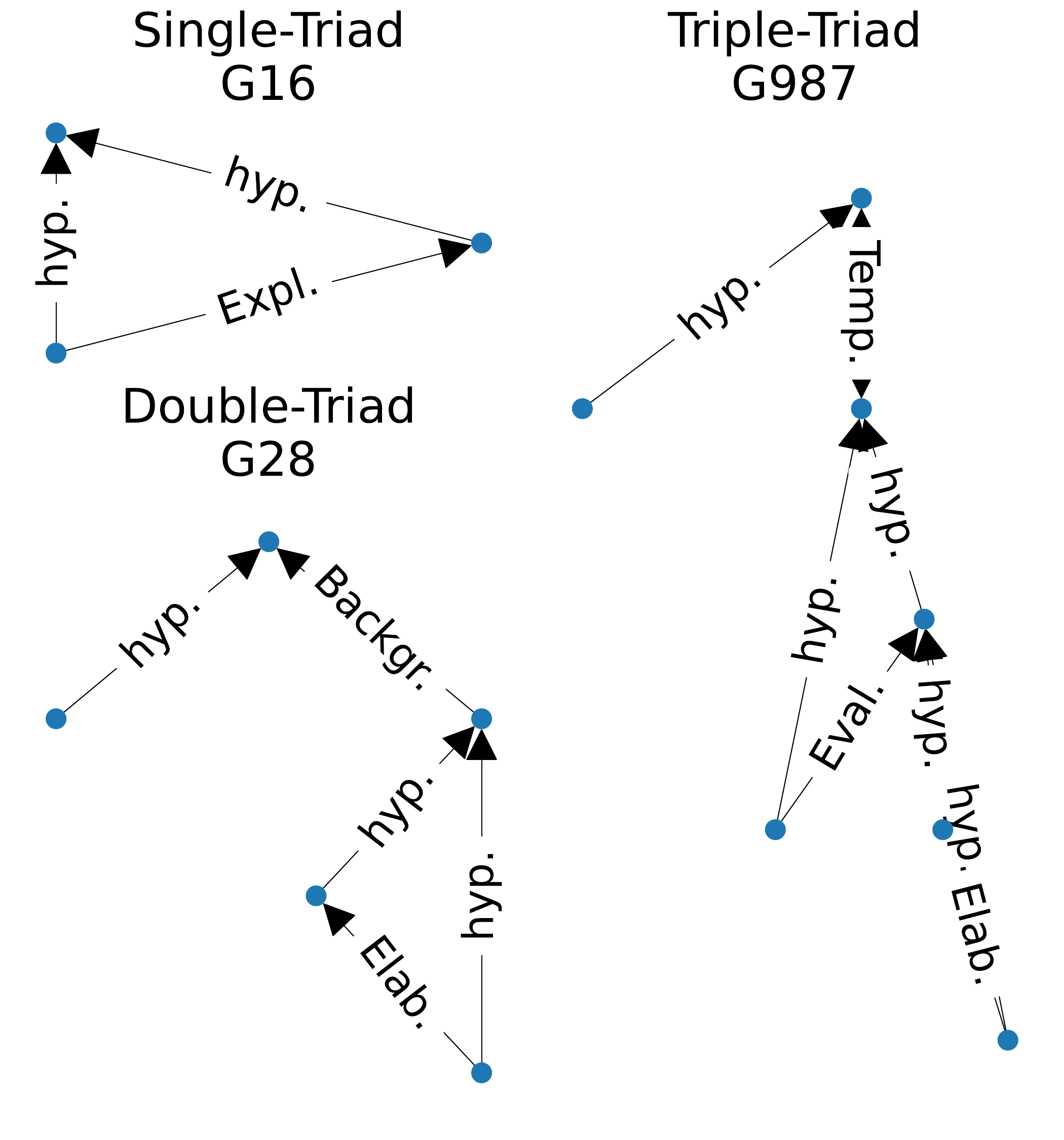}}
\caption{Examples of three types of motifs. More examples of single-triads are drawn in Figure \ref{fig:all_motifs_three}.\vspace{-4mm}}
\label{fig:3_6_9_motifs_examples}
\end{figure*}

\begin{figure*}[!ht]
\centering
{\includegraphics[width=\textwidth,height=0.9\textheight]{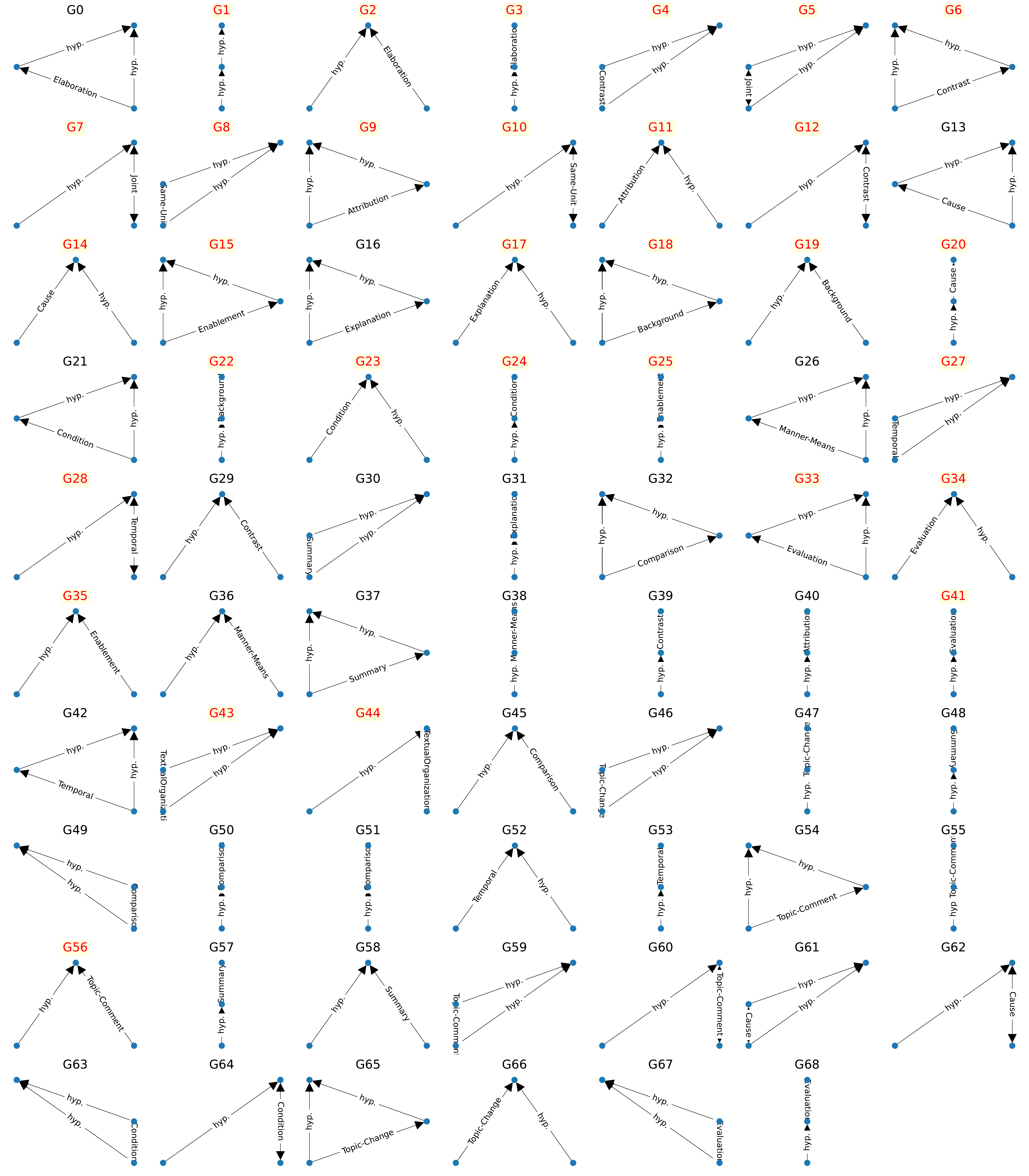}}
\caption{All 69 motifs of size three. The 31 marked motifs are the selected ones that exhibit MF-IDF scores surpassing at least one standard deviation.}
\label{fig:all_motifs_three}
\end{figure*}

Figure \ref{fig:3_6_9_motifs_examples} illustrates examples of the different types of triads. Similarly, Figure \ref{fig:all_motifs_three} shows all 69 single-triad motifs.

\end{document}